\documentclass[lettersize,journal]{IEEEtran}
\usepackage{amsmath,amsfonts}
\usepackage{algorithmic}
\usepackage{array}
\usepackage[caption=false,font=normalsize,labelfont=sf,textfont=sf]{subfig}
\usepackage{textcomp}
\usepackage{stfloats}
\usepackage{url}
\usepackage{verbatim}
\usepackage{graphicx}

\usepackage{multirow}
\usepackage{hyperref}
\usepackage{orcidlink}
\usepackage{amssymb}
\usepackage{booktabs}
\def\BibTeX{{\rm B\kern-.05em{\sc i\kern-.025em b}\kern-.08em
    T\kern-.1667em\lower.7ex\hbox{E}\kern-.125emX}}
\usepackage{balance}
\begin{document}
\title{Motif Channel Opened in a White-Box: Stereo Matching via\\Motif Correlation Graph}
\author{
Ziyang Chen\orcidlink{0000-0002-9361-0240},%~\IEEEmembership{Student Member,~IEEE,}
Yongjun Zhang*\orcidlink{0000-0002-7534-1219},~\IEEEmembership{Member,~IEEE,}
Wenting Li\orcidlink{0009-0009-8081-842X},
Bingshu Wang\orcidlink{0000-0002-2603-8328},\\
Yong Zhao\orcidlink{0000-0002-7999-1083},~\IEEEmembership{Member,~IEEE,}
C. L. Philip Chen\orcidlink{0000-0001-5451-7230},~\IEEEmembership{Life Fellow,~IEEE}% <-this % stops a space
\IEEEcompsocitemizethanks{
	\IEEEcompsocthanksitem Ziyang Chen and Yongjun Zhang are with the College of Computer Science, the State Key Laboratory of Public Big Data, Guizhou University, Guiyang 550025, China.
	\protect\\
	E-mail: ziyangchen2000@gmail.com, zyj6667@126.com
	\IEEEcompsocthanksitem Wenting Li is with the School of Information Engineering, Guizhou University of Commerce, Guiyang 550021, China.
	\protect\\
	E-mail: 201520274@gzcc.edu.cn
	\IEEEcompsocthanksitem Bingshu Wang is with the School of Software, Northwestern Polytechnical University, Xi'an, 710129, China.
	\protect\\
	E-mail: wangbingshu@nwpu.edu.cn
	\IEEEcompsocthanksitem Yong Zhao is with the School of Intelligent Manufacturing and Future Technology, Fujian Fuyao University of Science and Technology, Fuzhou 350109, China.
	\protect\\
	E-mail: zhaoyong@fyust.edu.cn
	\IEEEcompsocthanksitem C. L. Philip Chen is with the School of Computer Science and Engineering, South China University of Technology, Guangzhou 510641, China.
	\protect\\
	E-mail: philip.chen@ieee.org
}% <-this % stops an unwanted space
\thanks{* Corresponding author: Yongjun Zhang.}
%\thanks{Manuscript received November 19, 2024}%; revised August 26, 2015.}
}

\markboth{Journal of \LaTeX\ Class Files,~Vol.~18, No.~9, September~2020}%
{How to Use the IEEEtran \LaTeX \ Templates}

\maketitle

\begin{abstract}
Real-world stereo-matching applications demand high levels of accuracy. However, existing learning-based stereo matching methods are prone to losing geometric information across feature channels, limiting their ability to establish accurate correspondences for fine-grained details. In this paper, we present MoCha-V2, a novel learning-based stereo-matching paradigm. MoCha-V2 introduces a Motif Correlation Graph (MCG) to capture recurring texture patterns within individual feature channels, which we term ``motifs''. The MCG learns these motifs through a manually designed mechanism and uses them to reconstruct geometric structures. We then integrate features across multiple frequency bands using an inverse wavelet transform. The resulting motif features are used to recover geometric structures during stereo matching. Extensive experiments demonstrate the effectiveness of MoCha-V2. At the time of submission, MoCha-V2 ranked $1^{st}$ on the Middlebury benchmark.
%Code is available at
\href{https://github.com/ZYangChen/MoCha-Stereo}{Code}.
\end{abstract}


\section{Introduction}
\IEEEPARstart{S}{tereo} matching is a key component in computer vision applications, particularly in the fields of autonomous driving \cite{menze2015object,behl2017bounding,drivingstereo,navigation}, medical care \cite{med,li2021revisiting,bobrow2023colonoscopy}, 3D reconstruction \cite{tulyakov2017weakly,yao2018mvsnet,kaya2023multi}, and remote sensing \cite{bosch2019semantic,he2022hmsm,chen2024surface}. 
Creating a pixel-level disparity map is the main goal of binocular stereo matching \cite{lecunstereo}. This map can be used to calculate the depth of pixels in an image. The edge performance of the disparity map is particularly important for pixel-level rendering technologies such as virtual reality and augmented reality, where the exact match between scene model and image mapping is critical. This emphasises the importance of keeping the edges of the disparity map and the original RGB image precisely aligned. 

\begin{figure}[t]
	\centering
	\includegraphics[width=\linewidth]{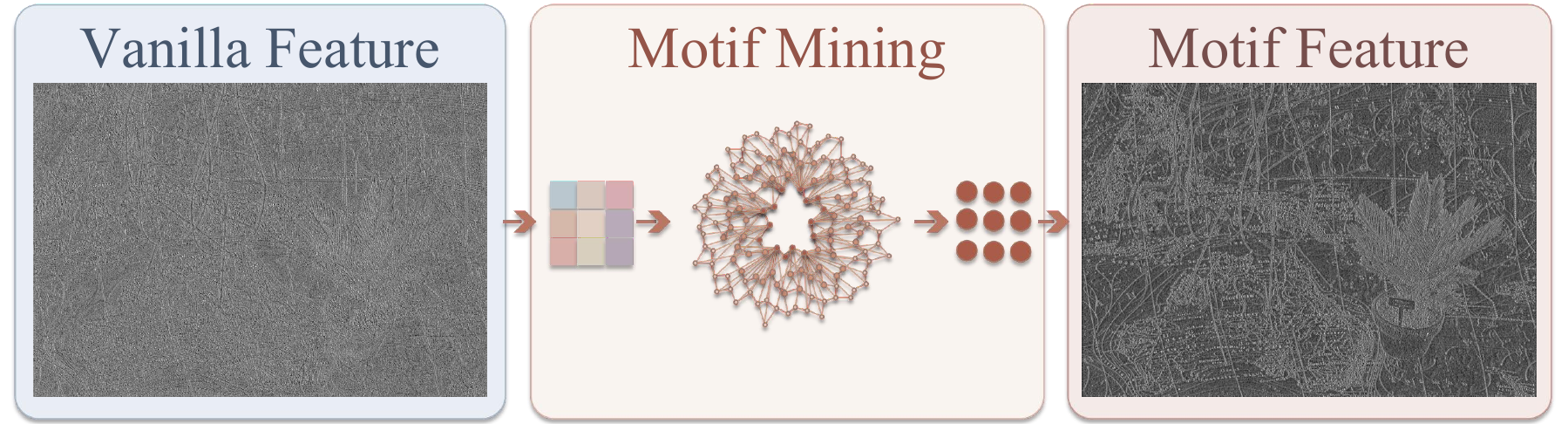}
	\caption{In the deep learning process, feature channels are extended to higher dimensions. During this process, the geometric details learned in individual feature channels are inevitably lost. We propose that these details can be restored by identifying recurrent geometric structures across the channels.  
	MoCha-V2 constructs a directed graph, i.e., motif correlation graph (MCG), to capture frequent patterns in feature channels. Specifically, MCG leverages node weights to detect recurring geometric structures within the channels. MCG establishes a robust and stable framework for identifying repetitive geometric structures. 
	}
	\label{motivation}
\end{figure}

Recent advances in deep learning have demonstrated the considerable potential of learning-based stereo matching methods \cite{simonyan2014very}. However, Deep Neural Networks (DNNs) map images from a low-dimensional input space to high-dimensional feature spaces. This transformation can produce uneven activations across feature channels, as observed in previous visualization studies \cite{visualizing,cam}. Consequently, some channels may lose fine-grained edge and texture information. Such feature abstraction can benefit image understanding by emphasizing high-level semantics. In stereo matching, however, accurate pixel-wise correspondence requires the precise preservation and matching of local details \cite{igev,edgestereo}. 

Building on these observations, our previous conference work introduced MoCha-Stereo \cite{mocha} and the concept of a ``motif channel.'' In time-series mining, a motif is a recurring subsequence \cite{mp}. 
By analogy, we define a motif channel as a feature channel that captures recurring geometric structures. We hypothesize that these structures can help recover edge information missing from vanilla channels. 
MoCha-Stereo uses sliding windows to extract recurring geometric patterns and projects the resulting motif information onto the corresponding feature channels. The enhanced features are then used to construct the correlation volume. Its state-of-the-art (SOTA) performance provides empirical support for recovering lost geometric details using recurring structures.

Despite the effectiveness of MoCha-Stereo, it still has several design limitations. \textbf{1) Black-box weight learning.} The motif-mining strategy in MoCha-Stereo uses learnable parameters to weight the elements within each sliding window. This black-box weighting mechanism may lead to unstable extraction of geometric structures, thereby compromising the robustness of fine-grained detail recovery across feature channels. 
\textbf{2) High-frequency patterns only.} While edge textures typically manifest as high-frequency information, recurrent low-frequency patterns also contributes to the understanding of details to some extent. 

To relieve these problems, we further design MoCha-V2. The key differences between MoCha-V2 and MoCha-Stereo are as follow.  \textbf{1) The recurring patterns are learned in a hand-crafted way.} The weights of sliding windows of MoCha-V2 are determined by the proposed Motif Correlation Graph (MCG). MCG is constructed based on the Euclidean distance between wavelet-domain feature channels. \textbf{2) MoCha-V2 further emphasises the importance of low-frequency information.} It further learns recurring low-frequency features in wavelet domain. 
With these improvements, MoCha-V2 ranked \textbf{$1^{st}$} on the Middlebury benchmark under the Bad 1.0 (all) metric at the time of submission.

In summary, our main contributions are as follow:

\begin{itemize}
	\item We propose MoCha-V2, a stereo matching framework that explicitly projects recurring geometric patterns across feature channels, enabling the recovery of fine-grained details and improving the accuracy and reliability of stereo correspondence.
	\item We introduce the Motif Correlation Graph (MCG) that mines recurring structural motifs from deep image features through a hand-crafted scheme. These motifs lead to discriminative deep features for robust matching.
	\item We design a feature restoration strategy that jointly exploits low-frequency and high-frequency information. By integrating these complementary frequency cues, our approach effectively repairs and enhances image features.
\end{itemize}

\section{Related Work}
Existing CNN-based stereo matching networks suffer from two critical
limitations: geometric detail is progressively lost in individual feature
channels~\cite{visualizing,cam}, and the process by which that detail is
recovered remains largely uninterpretable, which is a concern for
safety-critical applications such as autonomous driving and robotic
control. MoCha-V2 is designed against exactly these two limitations.

To make the position of our contribution explicit, we organize this
section along the lines of work that relate most closely to our two
design choices.
Section~\ref{sec:edge} reviews methods that recover edges and fine
details in the spatial domain; they share our objective, but locate the
cause of detail loss outside the feature channels.
Section~\ref{sec:freq} reviews methods that operate on frequency bands,
which is the domain in which MoCha-V2 searches for motifs.
Section~\ref{sec:attention} reviews attention mechanisms for stereo
matching and shows that existing designs have limitations when
transferred to this task directly, which motivates the proposed
Motif Correlation Graph Attention (MCGA) presented in
Section~\ref{sec:method}.

\subsection{Edge- and Detail-preserving Stereo Matching}
\label{sec:edge}

Preserving fine geometric detail is central to stereo matching, because
errors concentrate on edges, thin structures and textureless regions.
Existing methods pursue this goal from two directions. The first group
treats detail loss as a problem of insufficient supervision and
compensates by injecting external cues: SegStereo~\cite{segstereo}
leverages a segmentation network to provide semantic cues, and
EdgeStereo~\cite{edgestereo} employs an edge detection network to supply
edge cues, so that the extracted features are additionally constrained
towards semantically or structurally meaningful regions. The second group
treats it as a problem of inadequate representation and compensates by
enriching the volume or the features themselves:
IGEV-Stereo~\cite{igev} encodes geometric features into a combined
geometry encoding volume, and LoS~\cite{los} incorporates specialized
branches to capture local structure.

These methods and MoCha-V2 share the same objective, namely recovering fine geometric detail, but they differ in where the problem is located. The methods above situate it either in the supervision signal or in the cost and feature representation, and therefore add an auxiliary task, an auxiliary branch, or a richer volume.
MoCha-V2 instead situates it inside the CNN: because the network maps
features into a higher-dimensional space through non-linear
transformations, activation is imbalanced across channels and the
geometric detail carried by individual channels is
lost~\cite{visualizing,cam}. MoCha-V2 therefore introduces no auxiliary
task and no auxiliary branch; it reconstructs the missing detail from
recurring geometric structures that are already present among the
channels.

Interpretability is a further point of difference. In the methods
above---and in our own conference version~\cite{mocha}, whose
sliding-window weights are learned rather than derived---which detail is
recovered, and how, is determined by learned parameters and cannot be
traced back to a deterministic rule. In MoCha-V2, by contrast, the Motif
Correlation Graph (MCG) is built from Euclidean distances between
subsequences and can be inspected directly.

\subsection{Frequency-domain and High-frequency Methods}
\label{sec:freq}

Edges and fine textures are naturally described as high-frequency
components and smooth surfaces as low-frequency ones. A second line of
work therefore attacks the detail-loss problem by making the network
explicitly frequency-aware.

Outside stereo matching, FcaNet~\cite{fcanet} reformulates channel
attention as a frequency-analysis problem: it proves that global average
pooling is exactly the lowest-frequency component of the two-dimensional
discrete cosine transform, and replaces it with a multi-spectral
formulation that retains several frequency components. This is relevant
here because it shows that the information discarded when a feature
channel is squeezed into a scalar is precisely frequency information.

Within stereo matching, three groups of methods can be distinguished.
\textit{First}, methods that preserve high frequency through
architecture. DLNR~\cite{DLNR} observes that detail is progressively lost
during iterative updates, and addresses this with a channel-attention
transformer feature extractor, a multi-scale decoupled LSTM that
separates the hidden state used for disparity updates from the state
transferred across iterations, and a disparity-normalization refinement
module. High-frequency content is preserved implicitly, by designing
units that do not destroy it.
\textit{Second}, methods that operate on a frequency representation of
the disparity itself. WaveletStereo~\cite{waveletstereo} does not regress
the disparity map directly; it regresses its wavelet coefficients, one
sub-module producing the low-frequency approximation coefficients from
global context and the remaining sub-modules producing the high-frequency
detail coefficients.
\textit{Third}, methods that decompose the input into frequency bands and
process the bands separately. Selective-Stereo~\cite{selective} argues
that a recurrent unit with a fixed receptive field can only attend to a
single frequency band, and proposes a selective recurrent unit that fuses
hidden disparity states obtained with different receptive fields, the
fusion weights being produced by a contextual spatial attention module.

MoCha-V2 also performs an explicit two-level Haar discrete wavelet transform, but it uses the frequency domain for a different purpose and in a different way. 
1) \textit{Purpose.} The methods above use frequency bands either to
protect detail that would otherwise be destroyed during iteration \cite{DLNR,selective} or to provide the regressor with
a better-behaved target \cite{waveletstereo}. MoCha-V2 uses them as the space in which recurring structure is sought: a motif is defined over
subsequences of the wavelet-domain coefficients, because a recurring
geometric structure manifests itself as a recurring sequence of
coefficients.
2) \textit{Which bands.} Most of the above concentrate on the
high-frequency band, since edges reside there. MoCha-V2 mines motifs in
the low-frequency approximation subband as well: recurring low-frequency
patterns are part of what we call the motif channel, and
Section~\ref{sec:method} shows that they also contribute to detail
restoration.
3) \textit{How the bands are combined.} Existing methods combine bands with weights that are themselves learned.
In MoCha-V2, the motif weight of a node follows from the Euclidean
distance between subsequences, so the contribution
of each subband is determined by the data rather than by a learned
attention map. This is what makes the frequency reasoning of MoCha-V2
inspectable, whereas the frequency behaviour of existing methods \cite{DLNR,waveletstereo,selective} can only be observed after training.

\subsection{Attention Mechanisms in Stereo Matching}
\label{sec:attention}

Attention has been applied to stereo matching mainly through the three
standard forms: spatial attention, which reweights positions~\cite{stn}; channel attention, which reweights
channels~\cite{cama}; and mixed attention \cite{cbam,wu2024distribution}, which combines the two in series or in parallel. Mixed attention is motivated by the observation that the two forms pursue different goals: spatial attention is more concerned with edge details, and channel attention with semantic information. 

The properties of these modules make them difficult to transfer
directly to stereo matching. 
First, both detailed and semantic information are crucial for stereo matching, yet current mixed attention methods~\cite{cbam} treat the two aspects separately. The objectives of spatial and channel attention differ, so learning them in isolation compromises the joint capture of geometric details and global semantics. 
Second, existing attention methods do not consider the correlation
between the binocular views, which is central to stereo matching.

We propose Motif Correlation Graph Attention (MCGA) to
address the aforementioned issues. Spatial and channel features are
learned concurrently. The correlation between the left and right views is
modeled as a correlation graph. 

\section{Preliminaries}
\label{sec:preliminary}

\subsection{Motif From Time Series to Stereo Views}
\label{notion-motif}
``Motif'' was first defined in time series analysis and refers to a pattern that frequently appears within time series data. Mining motifs can be utilized to identify recurring patterns, detect anomalies, and perform pattern matching. Motifs are particularly useful for tasks such as time series classification and clustering \cite{matrix1,matrix2,matrix5}. The formal definition of a motif is as follows. 

In a time series $T$, there exists a subsequence $T_{i,L}$, which starts from the $i$-th position in the time series $T$ and is a continuous subset of values with a length of $L$. Motif is the pair of subsequences $T_{a,L}$ and $T_{b,L}$ in a time series that are the most similar. 
In mathematical notation, $\forall i,j \in [1,2,...,n-L+1]$ and $a \neq b\ , \ i \neq j$, the motif \cite{matrix2} satisfies as Equation \ref{motif}.
\begin{gather}
	\label{motif}
	\begin{aligned}
		\centering
		dist(T_{a,L},T_{b,L}) &\leq dist(T_{i,L},T_{j,L}) 
	\end{aligned}
\end{gather}
where $dist$ means a distance measure. The distances between all subsequences and their nearest neighbors are stored as Matrix Profile (MP) \cite{matrix1}, representing the repetitive pattern (motif) in the time series. 

In stereo matching, the geometric structure also repeats across binocular views. However, the computation of motif patterns (MP) in time series is not directly applicable to stereo matching. Unlike time series, the repetitive patterns (or motifs) in images are two-dimensional. The MP computation in time series is computationally expensive for images, as it requires multiple samplings of subsequences. Additionally, selecting sub-patches from multi-channel image features to compute similarity incurs significant computational cost.

To address this problem, we have developed a collection of sliding windows, which are arranged into two-dimensional vectors to capture recurring geometric patterns. These patterns are then stored as ``motif channels''.

\subsection{Iterative Update Operator}
\label{3.5}
Since RAFT-Stereo~\cite{raftstereo} introduced a convolutional gated recurrent unit (ConvGRU) as the iterative update operator for disparity
refinement, this recurrent paradigm has been widely adopted by subsequent
learning-based stereo methods. 
DLNR~\cite{DLNR} notes that, in a GRU-based operator, a single hidden state is hard to simultaneously support the regression of the disparity update and the transfer of information across iterations. To alleviate this problem, DLNR replaces the GRU with a long short-term memory unit implemented by convolutions, a ConvLSTM, which decouples the two roles through an explicit cell state. 
Both MoCha-Stereo \cite{mocha} and the proposed MoCha-V2 follow this ConvLSTM-based paradigm.

Formally, let $\mathbf{C}$ denote the correlation volume and $\mathbf{d}_{t-1}$ the disparity estimate obtained at iteration $t-1$. At iteration $t$, the update operator first performs a lookup on $\mathbf{C}$ around $\mathbf{d}_{t-1}$ to gather the local matching costs, and concatenates them with the context features and the current disparity itself to form the operator input
\begin{equation}
	\mathbf{x}_{t}
	= \big[\, \mathcal{L}_{\mathbf{C}}(\mathbf{d}_{t-1}),\;
	\mathbf{f}_{\mathrm{ctx}},\;
	\mathbf{d}_{t-1} \,\big],
\end{equation}
where $\mathcal{L}_{\mathbf{C}}(\cdot)$ denotes the correlation lookup and
$\mathbf{f}_{\mathrm{ctx}}$ is the context feature map. The ConvLSTM then
updates its states according to
\begin{align}
	\mathbf{i}_{t} &= \sigma\big( \mathbf{W}_{xi} * \mathbf{x}_{t}
	+ \mathbf{W}_{hi} * \mathbf{h}_{t-1} + \mathbf{b}_{i} \big),
	\\[2pt]
	\mathbf{f}_{t} &= \sigma\big( \mathbf{W}_{xf} * \mathbf{x}_{t}
	+ \mathbf{W}_{hf} * \mathbf{h}_{t-1} + \mathbf{b}_{f} \big),
	\\[2pt]
	\mathbf{o}_{t} &= \sigma\big( \mathbf{W}_{xo} * \mathbf{x}_{t}
	+ \mathbf{W}_{ho} * \mathbf{h}_{t-1} + \mathbf{b}_{o} \big),
	\\[2pt]
	\mathbf{g}_{t} &= \tanh\big( \mathbf{W}_{xg} * \mathbf{x}_{t}
	+ \mathbf{W}_{hg} * \mathbf{h}_{t-1} + \mathbf{b}_{g} \big),
	\\[2pt]
	\mathbf{c}_{t} &= \mathbf{f}_{t} \cdot \mathbf{c}_{t-1}
	+ \mathbf{i}_{t} \cdot \mathbf{g}_{t},
	\\[2pt]
	\mathbf{h}_{t} &= \mathbf{o}_{t} \cdot \tanh\big( \mathbf{c}_{t} \big),
\end{align}
where $*$ denotes the two-dimensional convolution, $\cdot$ the Hadamard
product, $\sigma(\cdot)$ the sigmoid function, $\mathbf{i}_{t}$,
$\mathbf{f}_{t}$ and $\mathbf{o}_{t}$ the input, forget and output gates,
$\mathbf{c}_{t}$ the cell state, and $\mathbf{h}_{t}$ the hidden state. Since
all gating operations are realized by convolutions rather than fully connected
layers, the spatial layout of the states is preserved throughout the
iterations, which allows the operator to refine the disparity field densely
over the image. 
Given the updated hidden state, the operator regresses a disparity residual
and applies it to the current estimate as
\begin{align}
	\Delta\mathbf{d}_{t} &= \mathcal{R}\big( \mathbf{h}_{t} \big), \\
	\mathbf{d}_{t} &= \mathbf{d}_{t-1} + \Delta\mathbf{d}_{t},
\end{align}
where $\mathcal{R}(\cdot)$ is a lightweight convolutional head. It is the formulation adopted by MoCha-Stereo \cite{mocha} and MoCha-V2.

\begin{figure*}[t]
	\centering
	\includegraphics[width=\linewidth]{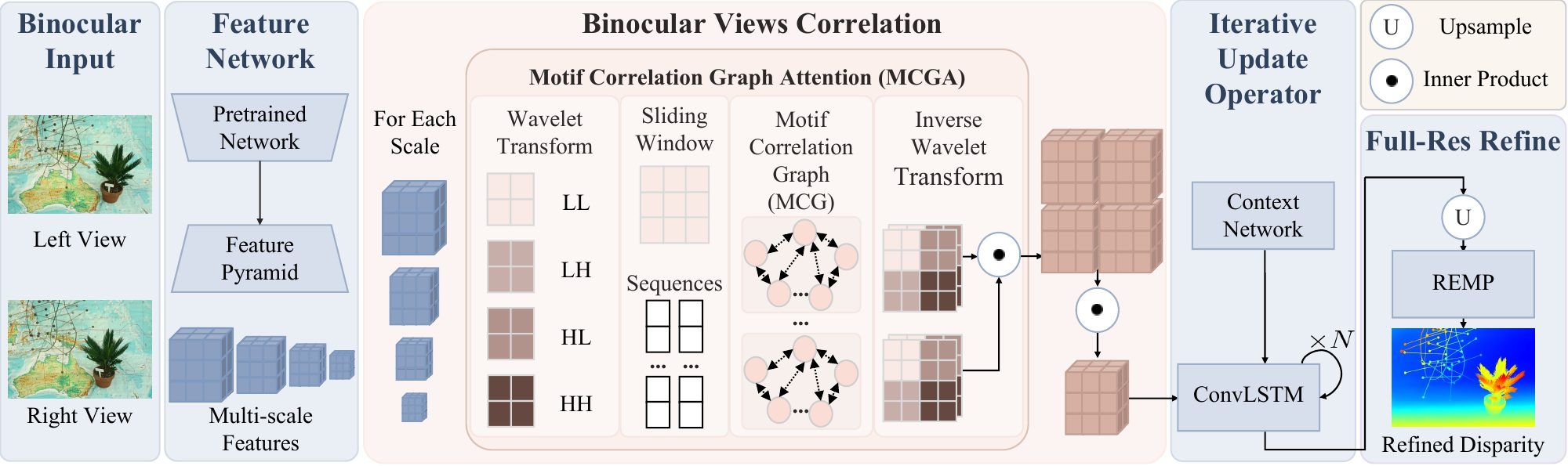}
	\caption{Architecture overview of MoCha-V2. First, a feature network extracts multi-scale features from the input stereo pair, and a channel-wise Haar wavelet transform decomposes the features into four subbands. A sliding window partitions each subband into non-overlapping \(3\times3\) patches, which are treated as candidate motif subsequences. Second, for each channel group and window location, we construct a Motif Correlation Graph (MCG) in which nodes are these subsequences and directed edges are determined by Euclidean nearest-neighbour relationships; each node casts one unit vote to its nearest neighbours, and the vote-weighted mean of subsequences gives the motif maps. After applying the inverse wavelet transform and a sigmoid activation, each group yields a single motif channel. Third, the motif channels are broadcast and element-wise multiplied with the original vanilla channels, so that non-recurring patterns are attenuated while recurring geometric details are retained; this restoration step is parameter-free and thus fully interpretable. Fourth, the restored features are used to build a motif-weighted group-wise correlation volume, from which a differentiable soft argmin regresses an initial disparity map. The initial disparity is then refined by an iterative ConvLSTM update operator and further corrected by the Reconstruction Error Motif Penalty (REMP) module. 
	} 
	\label{pipeline}
\end{figure*}

\subsection{Reconstruction Error Motif Penalty (REMP)}
\label{3.6}

We refine the upsampled disparity map $\mathbf{d}_n$ using the Reconstruction Error Motif Penalty (REMP) module, inherited unchanged from our conference version \cite{mocha}. It computes a warping-based reconstruction error $\mathbf{E}$ from the camera parameters, and feeds it into a UNet with three branches: a Low-Frequency Error (LFE) branch that uses pooling to preserve semantic structure, a learned mask branch producing a spatial attention mask $\mathbf{M}(\mathbf{o})$, and a branch computing $\mathbf{o} \odot \mathbf{M}(\mathbf{o})$ to retain informative latent features. The final refinement combines these branches to correct $\mathbf{d}_n$.

\section{Method}
\label{sec:method}

\subsection{Overview}
\label{sec:overview}

In this section, we introduce the overall framework of MoCha-V2, as illustrated in Figure \ref{pipeline}. The key idea is to recover fine-grained geometric details that are unevenly lost across feature channels by exploiting recurring geometric structures, i.e., motifs, in both high- and low-frequency domains. In the following, we describe each component in order.

\subsection{Feature Network}
\label{3.2}
Given a stereo image pair, a feature network is utilized to extract multi-scale features, following \cite{mocha}. The feature network uses a backbone pretrained on ImageNet \cite{deng2009imagenet} as the frozen layer \cite{tan2019efficientnet}. The upsampling blocks utilizes skip connections from outputs of downsampling blocks to obtain multi-scale outputs $\mathbf{F}_{l(r)}{ \in \mathbb{R}^{N_c \times \frac{H}{i} \times \frac{W}{i}} }$ ($i=4, 8, 16, 32$), $i$ means scale, $N_c$ represents the feature channels, and $\mathbf{F}_{l(r)}$ denotes the left (right) view features here. 
We call each of the $N_c$ channels of $\mathbf{F}_{l(r)}$ a vanilla channel, i.e., a feature channel that has not been modulated by a motif.

\subsection{Motif Extraction}
The extracted repeated patterns are preserved as ``motifs'', as illustrated in Section \ref{notion-motif}. As in our previous work, MoCha-Stereo \cite{mocha} extracts motifs in the
high-frequency domain. However, a high-pass filter alone limits our ability to learn repeated semantic patterns in the low-frequency domain. 

To capture motifs in both high- and low-frequency domains, we employ a Haar discrete wavelet transform \cite{haar,parida2017wavelet}, which decomposes the input feature into four subbands: LL (Low-Low, i.e. low-frequency along both axes), LH (Low-High), HL (High-Low) and HH (High-High). LL is the approximation subband and reflects the low-frequency components, whereas LH, HL and HH are detail subbands that represent the high-frequency components along the two axes and along the diagonal, offering high-frequency values for both the foreground and the background. The transform is applied channel-wise, so for an input feature $\mathbf{F} \in \mathbb{R}^{N_c \times \frac{H}{4} \times \frac{W}{4}}$ each subband retains all $N_c$ channels and has half spatial resolution, i.e. size $N_c \times \frac{H}{8} \times \frac{W}{8}$. The $N_c$ channels are partitioned into $N_g$ disjoint groups, each of size $N_c/N_g$. 

Subsequently, we use a sliding window (SW) to capture motifs. The original sliding window in MoCha-Stereo \cite{mocha} uses four sets of adaptive weights to map motifs. However, learning these weights with a convolutional neural network makes it difficult to explain how they are generated. This causes the learning of recurrent geometric structures to remain unstable in certain motif channels. 
In this version, $SW$ is only used to intercept segments on the feature channel without additional manipulation. $SW$ slides a $3 \times 3$ window over each channel with stride $3$, thereby partitioning each channel into $k$ non-overlapping patches, and flattening the $j$-th patch of the $c$-th channel into a one-dimensional sequence $s_{c,j} \in \mathbb{R}^{9}$ ($1 \leq c \leq N_c$, $1 \leq j \leq k$). That is, each feature channel is transformed into $k$ sequences, each of length $9$. %

\subsection{MCG Construction}
Since the $N_c$ channels have been partitioned into $N_g$ groups, each group thus contains $(N_c/N_g)\cdot k$ sequences. We regard each group as one graph in which every sequence is a node. A node $n_{c,j}$ is the subsequence $s_{c,j}$, i.e. the $3\times3$ patch at location $j$ of channel $c$ of group $g$, flattened into a $9$-D vector. It is therefore a candidate motif instance; its value is that $9$-D subsequence and its weight $w_{c,j}$ is a scalar measuring how representative it is. s
These obtained segments are used to construct the Motif Correlation Graph (MCG).

Because a motif is by definition a local recurring pattern, edges are
formed only among nodes at the same window location $j$; the subgraph induced by the $N_c/N_g$ nodes at location $j$ is denoted $\mathcal{G}_{g,j}$, and there are $k \cdot N_g$ such subgraphs per subband, one per group and window location. The graph is directed: each node casts exactly one unit of vote, so it emits a directed edge to each of its nearest neighbours $\bar{c} \neq c$ under the Euclidean distance $d(s_{c,j}, s_{\bar{c},j}) = \| s_{c,j} - s_{\bar{c},j} \|_2$, the self-pair $d(s_{c,j}, s_{c,j}) = 0$ being excluded, and that edge carries the amount of vote transferred. The distance is thus the criterion that determines which directed edges exist, while the edge weight is the vote itself. This graph is genuinely directed: that $c$ votes for $\bar{c}$ does not imply that $\bar{c}$ votes for $c$. Consequently every node has unit out-weight (split as $1/p_{c,j}$ among $p_{c,j}$ tied nearest neighbours), while the weighted in-degree of node $n_{c,j}$ is exactly its weight $w_{c,j}$: how much a node is voted for is precisely how representative it is. The above construction is applied to each of the four subbands independently, so that motifs are extracted within each frequency band.

\subsection{From Graphs to Motif Channels}
Within each subgraph $\mathcal{G}_{g,j}$, each node casts exactly one unit of vote for the node(s) closest to it. Let $d^{\star}_{c,j} = \min_{\bar{c} \neq c} d(s_{c,j}, s_{\bar{c},j})$ be the minimum value and $\mathcal{N}_{c,j} = \{ \bar{c} \neq c \mid d(s_{c,j}, s_{\bar{c},j}) =d^{\star}_{c,j} \}$ its argument set, with $p_{c,j} = |\mathcal{N}_{c,j}|$. Then $w_{\bar{c},j}$ is incremented by $1/p_{c,j}$ for every $\bar{c} \in \mathcal{N}_{c,j}$. The minimum value
$d^{\star}_{c,j}$ is unique, whereas its argument may contain $p_{c,j} \geq 1$ elements; the factor $1/p_{c,j}$ distributes a single voter's unit vote equally among tied candidates, so that the result does not depend on channel ordering. Because each of the $N_c/N_g$ nodes of $\mathcal{G}_{g,j}$ casts exactly one vote whether or not it is involved in
a tie, it follows that $\sum_c w^{g}_{c,j} = N_c/N_g$ for every group $g$ and every window location $j$. 

$\mathcal{G}_{g,j}$ is the weighted mean of its candidate subsequences under this vote distribution:
\begin{equation}
	\label{eq:motif}
	m_{g,j} =
	\frac{N_g}{N_c} \sum_{c=1}^{N_c/N_g} w^{g}_{c,j}\; s^{g}_{c,j},
	\quad 1 \leq g \leq N_g,\;\; 1 \leq j \leq k,
\end{equation}
where the factor $N_g/N_c$ normalises the weights ($\sum_c w^{g}_{c,j} = N_c/N_g$) so that Eq.~\eqref{eq:motif} is an average and hence remains on the same scale as its inputs. 
$w_{c,j}$ counts how many other subsequences at the same location regard
$s_{c,j}$ as their nearest neighbour. A large $w_{c,j}$ is therefore
desirable: it indicates that $s_{c,j}$ is a representative and frequently
recurring pattern. A weight of zero means the subsequence is an outlier that
no other subsequence recurs to, and it is discarded from the motif. Writing
$\pi_{c,j} = \frac{N_g}{N_c} w^{g}_{c,j} \in [0,1)$ for the vote share of
node $n_{c,j}$, Eq.~\eqref{eq:motif} is simply the weighted mean of the
candidate subsequences with respect to this distribution: MCG is a
parameter-free, vote-weighted mean, that suppresses non-recurring patterns. 

The $k$ motifs $\{m_{g,j}\}_{j=1}^{k}$ that group $g$ obtains on one subband, each of length $9$, are reshaped back into $3\times3$ blocks and stitched at their window positions, which restores a motif map at subband resolution. Every
$s_{c,j}$ is itself a flattened $3\times3$ patch, so the weighted mean of
Eq.~\eqref{eq:motif} is the flattening of the mean patch, and no information is
lost. Because the windows are non-overlapping, the reconstruction is exact: the
$k$ blocks tile the motif map with neither overlap nor holes. Repeating this on
all four subbands yields the four motif maps that the inverse transform expects;
we collect them into
$M^{g} = \{ M^{g}_{\mathrm{LL}}, M^{g}_{\mathrm{LH}}, M^{g}_{\mathrm{HL}},
M^{g}_{\mathrm{HH}} \}$. The inverse wavelet transform (IWT) takes these four
maps as its input and returns, for each group, a single motif channel
\begin{equation}
	G^{g} = \sigma\!\big(\mathrm{IWT}(M^{g})\big)\in\mathbb{R}^{H/4\times W/4},
\end{equation}
\noindent
which has exactly the spatial resolution of $\mathbf{F}_{l(r),4}$.

\subsection{Vanilla Channel Restoration}
We restore vanilla channels with motif channels. Recall that the $N_c$ vanilla feature channels are partitioned into $N_g$ groups of equal size $\hat C=N_c/N_g$, the $g$-th group owning the channels $c\in\{(g-1)\hat C+1,\ldots,g\hat C\}$, which we index by $g(c)=\lceil c/\hat C\rceil$. The single motif channel $G^{g}$ produced by group $g$ is shared by all the channels of that group: it is broadcast along the channel dimension, i.e.\ replicated $\hat C$ times so as to match the $N_c$ channels of $\mathbf{F}_{l(r),4}$. In element-wise form, with $(x,y)$ the coordinates of a pixel in the $c$-th channel,
\begin{gather}
	\mathbf{F}^{\mathrm{mo}}_{l(r),4}(c,x,y)
	= \mathbf{F}_{l(r),4}(c,x,y)\odot G^{g(c)}_{x,y}
	\label{final_motif}
\end{gather}
\noindent
where $c=1,\ldots,N_c$, $x=1,\ldots,H/4$ and $y=1,\ldots,W/4$. Thus the $c$-th vanilla channel is modulated by the motif channel of its own group and by no other. Here $\mathrm{IWT}$ denotes the inverse wavelet transform, $\sigma$ is
the element-wise logistic sigmoid and $\odot$ denotes element-wise
multiplication. Haar detail coefficients are signed differences, so
$\mathrm{IWT}(M^{g})$ may in principle take negative values; the sigmoid
confines the gate to $(0,1)$, so that it can never flip the sign of a feature
and acts purely as a multiplicative attenuation: a value near $0$ fully
suppresses a location that carries no recurring pattern, while a value near $1$
leaves the feature untouched. Since the broadcast is determined by the fixed
grouping $g(c)$ and $\sigma$ contains no learnable parameter, the module remains
fully white-box.

The MCG output is detached from the computation graph before
Eq.~\eqref{final_motif} is applied. This is not an approximation: because
the votes are produced by an $\arg\min$, every $w^{g}_{c,j}$ is a
piecewise-constant function of the subsequences, the same situation as the $\arg\max$ of max-pooling. Gradients of the disparity loss therefore flow back through $f(c,x,y)$ to the feature extractor, which remains trainable in an end-to-end manner, while MCG serves as a fixed, data-dependent geometric prior.

In this way, our MCG implements both forms of channel attention and spatial attention, i.e., Motif Correlation Graph Attention (MCGA). Along the channel axis, Eq.~\eqref{eq:motif} aggregates the patch
sequences with the normalized votes, which is the operation of a channel attention module~\cite{cama}, with the difference that the weight is conditioned on the spatial patch rather than being a single global scalar per channel; we therefore refer to it as a locally-conditioned channel reweighting. Along the spatial axes, the transformed motif map $\mathrm{IWT}(M^{g})$ is applied element-wise in Eq.~\eqref{final_motif}, attenuating those locations whose local pattern does not recur across channels, which is the form of a spatial attention module~\cite{stn}. The $3\times3$ window plays the role of the pooling neighborhood that defines this spatial support, with the difference that our statistics are obtained in a white-box manner from the Euclidean distances rather than learned by a convolution. 
By this means, we achieve spatial and channel attention simultaneously, rather than modeling them sequentially as in mixed-attention approaches such as CBAM \cite{cbam}.

\subsection{Binocular Views Correlation for Disparity Calculation}
\label{3.4}

Given the motif-restored features $\mathbf{F}^{\mathrm{mo}}_{l,4}$ and
$\mathbf{F}^{\mathrm{mo}}_{r,4}$ of Eq.~\eqref{final_motif}, we construct a
motif-weighted group-wise correlation volume and regress the disparity from it.
Let $\mathbf{f}^{[g]}_{l(r),4}(x,y)\in\mathbb{R}^{\hat C}$ be the slice of the
restored left (right) feature at position $(x,y)$ that belongs to group $g$,
i.e.\ the entries $c$ with $g(c)=g$, where $g(c)=\lceil c/\hat C\rceil$ and
$\hat C=N_c/N_g$. The group-wise inner-product correlation is
\begin{equation}
	\mathcal{C}_g(d,x,y)=\frac{1}{\hat C}\,
	\big\langle \mathbf{f}^{[g]}_{l,4}(x,y),\,
	\mathbf{f}^{[g]}_{r,4}(x-d,y)\big\rangle,
	\quad d=0,\ldots,D-1,
	\label{eq:group_corr}
\end{equation}
where $D$ is the number of discrete disparity candidates at $1/4$ resolution and
out-of-range positions are zero-padded. Because the motif channel $G^{g}$ of
Eq.~\eqref{final_motif} is computed for the two views separately, denoted
$G^{g}_{l}$ and $G^{g}_{r}$, it also provides a parameter-free measure of how
well the recurring structures of the two views agree at a candidate $d$. We use
their product as a motif similarity weight,
\begin{equation}
	\omega_g(d,x,y)=G^{g}_{l}(x,y)\,G^{g}_{r}(x-d,y)\in(0,1),
	\label{eq:motif_weight}
\end{equation}
\noindent
and obtain the motif correlation volume by reweighting and summing the $N_g$
group volumes,
\begin{equation}
	\mathcal{C}(d,x,y)=\sum_{g=1}^{N_g}\omega_g(d,x,y)\,\mathcal{C}_g(d,x,y).
	\label{eq:mccv}
\end{equation}
\noindent
The weight suppresses a candidate whose recurring structure is absent in either
view, while leaving the group correlation untouched where both views exhibit the
same motif; no learnable parameter is involved.

An initial disparity is then regressed from $\mathcal{C}$ by the differentiable
soft argmin
\begin{equation}
	d_0(x,y)=\sum_{d=0}^{D-1} d\cdot \mathrm{softmax}_{d}\big(\mathcal{C}(d,x,y)\big),
	\label{eq:soft_argmin}
\end{equation}
\noindent
which is bilinear upsampled to full resolution. The initial disparity is
subsequently refined by the iterative update operator described in
Section~\ref{3.5}, and the resulting map is further corrected at full
resolution by the REMP module described in Section~\ref{3.6}, yielding the
final disparity $\mathbf{d_n}$.

\begin{table}[b]
	\centering
	\caption{Training and testing protocols used for each result table. 
		For every benchmark only its official training split is used for fine-tuning, and the test split used for evaluation is never seen during training.}
	\label{tab:protocol}
	\begin{tabular}{ccc}
		\toprule[1.5pt]
		Table              & Test benchmark (split)        & Training set              \\ \hline
		\ref{SF},\ref{abl}-\ref{iter_time}             & SceneFlow (test)              & S1: SF                    \\ \midrule
		\multirow{2}{*}{\ref{mid2014}} & Middlebury V3                 & S1: SF+TA+CR+FT+IS2+CH+MB \\
		& (test)                        & S2: CR+FT+IS2+CH+MB       \\ \midrule
		\multirow{2}{*}{\ref{reflective2012}} & KITTI 2012                    & S1: SF                    \\
		& (reflective test)             & S2: K15+K12               \\ \midrule
		\multirow{2}{*}{\ref{kittitable}} & KITTI 2012                    & S1: SF                    \\
		& +2015 (test)                  & S2: K15+K12               \\ \midrule
		\multirow{2}{*}{\ref{eth3d_tab}} & \multirow{2}{*}{ETH3D (test)} & S1: TA+CR+SF+ST+IS2+E3    \\
		&                               & S2: CR+IS2+E3            \\
		\bottomrule[1.5pt]
	\end{tabular}
	\begin{minipage}{\columnwidth}
		S1 and S2 denote the first and second fine-tuning stage, respectively;
		a single S1 entry means no second stage is used.\\
		\textbf{Abbreviations:} SF = SceneFlow; MB = Middlebury Version 3 (train)+Middlebury 2005+Middlebury 2006+Middlebury 2014+Middlebury 2021; K15 = KITTI 2015 (train);
		K12 = KITTI 2012 (train); E3 = ETH3D (train); CR = CREStereo; TA = TartanAir;
		FT = FallingThings; IS2 = InStereo2k; CH = CARLA HR-VS; ST = Sintel.
	\end{minipage}
\end{table}

\subsection{Loss Function}
\label{3.7}
Following \cite{igev,mocha}, the computation of the loss function requires the disparity maps outputted at each iteration, and the initial disparity map $\mathbf{d_0}$. 
The total loss is defined as
\begin{gather}
	L = \mathit{Smooth}_{L1}( d_0-d_{gt} ) + \sum_{k=1}^{n} \gamma^{n-k} \Vert d_k - d_{gt} \Vert _{1}
	\label{losst}
\end{gather}
where $\mathit{Smooth}_{L1}$ is defined by \cite{r-cnn}, $\gamma = 0.9$, $d_{gt}$ is the ground truth disparity.

\section{Experiments}
\subsection{Datasets and Evaluation Metrics}
To evaluate the performance of our model, we conducted assessments using the Scene Flow \cite{sceneflow} datasets. In order to validate the performance of MoCha-V2 in real-world scenarios, we conducted experiments on the KITTI 2012 \cite{kitti2012}, KITTI 2015 \cite{kitti2015}, Middlebury \cite{middlebury}, and ETH3D \cite{eth3d} datasets.

\textbf{Scene Flow} \cite{sceneflow}. There are 4,370 picture pairs for testing and 35,454 for training in this sizable synthetic dataset. Each image has a resolution of 540 $\times$ 960 pixels and includes a dense ground-truth disparity for training. 
The evaluation metrics include discrepancy outlier rate Bad 1.0 and end-point error (EPE). Pixels with disparity errors more than 1px are referred to as the Bad 1.0.

\begin{table}[h]
	\centering
	\caption{Quantitative evaluation on Scene Flow test set. The best result is \textbf{bolded}, and the second-best result is \underline{underlined}. ``EPE" means end-point error. ``D1\textgreater1px" means percentage of stereo disparity outliers in first frame, and the error threshold is 1px.}
	\label{SF}
\begin{tabular}{l|cc}
	\toprule[1.5pt]
	Method      & EPE (px) & D1\textgreater1px (\%) \\	\midrule
	PSMNet \cite{psmnet}    &     1.09    &     12.1    \\
	RAFT-Stereo \cite{raftstereo}  & 0.60 &  6.5 \\
	DLNR \cite{DLNR}  & 0.48     &     5.4    \\
	IGEV-Stereo \cite{igev}      &     0.47    &    5.3     \\
	MC-Stereo \cite{mc} &     0.45    &     \underline{5.0}    \\
	%HART \cite{hart}       &    0.42    &    \underline{5.0}    \\	
	Selective-IGEV \cite{selective}   &    0.44    &    \underline{5.0}    \\	
	GEAStereo \cite{geastereo} &    \underline{0.40}   &   -  \\	
	GGEV \cite{ggev} &  0.45   &   -  \\	
	\midrule
	MoCha-Stereo \cite{mocha}   & \multirow{2}{*}{0.41}     &    \multirow{2}{*}{\underline{5.0}}     \\
	(conference version)& & \\
	MoCha-V2 (Ours)   &   \textbf{0.38}   &   \textbf{4.3}     \\
	\bottomrule[1.5pt]
\end{tabular}
\vspace{1pt}
\caption{Results on the Middlebury benchmark at the time of submission. For all metrics in this table, lower values indicate better performance. ``D1" means percentage of stereo disparity outliers in first frame. ``Avg" refers to the average D1 across all pixels in the Middlebury test sets. ``Rank" denotes to the ranking of methods on the Middlebury benchmark, which includes both published and unpublished methods. \textbf{Bold}: best performance, \underline{underline}: second best.}
\label{mid2014}
\begin{tabular}{l|cccc|ccc}
	\toprule[1.5pt]
	\multirow{2}{*}{Method} &  & \multicolumn{2}{c}{D1\textgreater0.5px (\%) $\downarrow$} &  &  & \multicolumn{2}{c}{D1\textgreater1px (\%) $\downarrow$} \\ \cline{3-4} \cline{7-8} 
	&  & Avg & Rank &  &  & Avg & Rank \\	\midrule
	RAFT-Stereo \cite{raftstereo} &  & 33.0 & 10 &  &  & 15.1 & 17 \\
	CREStereo++ \cite{cre++} &  & 33.3 & 14 &  &  & 16.5 & 20 \\
	CroCo-Stereo \cite{croco}  &  & 44.4 & 35 &  &  & 21.6 & 38 \\
	IGEV-Stereo \cite{igev}  &  & 36.6 & 15 &  &  & 13.8 & 13 \\
	UCFNet \cite{ucfnet}  &  & 55.6 & 98 &  &  & 31.6 & 96 \\
	GMStereo \cite{gmstereo}  &  & 54.8 & 91 &  &  & 28.4 & 82 \\
	LoS \cite{los} &  & 35.1 & 13 &  &  & 14.2 & 15 \\
	%HART \cite{hart} & arXiv2025 &  & 29.8 & 4 &  &  & 12.6 & 5 \\
	ADStereo \cite{wang2025adstereo}  &  & 63.8 & $\textgreater$ 100 &  &  & 39.9 & $\textgreater$ 100 \\
	Selective-IGEV \cite{selective} &  & \underline{29.6} & \underline{3} &  &  & \underline{11.4} &  \underline{2}\\	
	\midrule
	MoCha-V2 (Ours)  &  & \textbf{29.4} & \textbf{2} &  &  & \textbf{11.4} & \textbf{1} \\
	\bottomrule[1.5pt]
\end{tabular}
\end{table}

\textbf{Middlebury} \cite{middlebury} is an indoor stereo benchmark captured with structured light in real indoor scenes. 
The Middlebury Version 3 benchmark provides 15 training pairs and 15 test pairs with densely labeled ground truth; each scene is distributed at full (F), half (H), and quarter (Q) resolutions. 
Besides the Version 3 test set, it releases several historical and supplementary collections, namely Middlebury 2005, 2006, 2014, and 2021. 
When training, we use the Version3 training split together with licensed 2005, 2006, 2014, 2021 pairs for supervision, and report final numbers on the Version3 hidden test set. 
The evaluation metric, Bad 1.0, is defined as the percentage of pixels with an EPE greater than one pixels.

\textbf{KITTI 2012} \cite{kitti2012} and \textbf{KITTI 2015} \cite{kitti2015} consist of images taken in real-world outdoor driving environments. LIDAR-derived sparse ground-truth disparities are provided in both datasets. KITTI 2012 contains 389 image pairs, with 194 used for training and 195 for testing. In the KITTI 2015 dataset, there are 200 pairs for testing and 200 pairs for training. As the validation set, we divided 20 training pairings in accordance with \cite{igev,gwc}. Only the training set ground truth is accessible for both KITTI 2012 and 2015, and test results must be submitted to the benchmark for evaluation and ranking. For KITTI 2012, we present the proportion of pixels with errors greater than $3$ disparities in both non-occluded (Out-noc) and all regions (Out-all). For KITTI 2015, the percentage of pixels in the background (D1-bg), foreground (D1-fg), and all (D1-all) areas with EPE greater than three pixels is reported.

\textbf{ETH3D} \cite{eth3d} is a dataset that contains both indoor and outdoor scenes, with 20 low-resolution two-view image pairs for testing and 27 for training.

%\textbf{Driving Stereo} \cite{drivingstereo} contains 2,000 frames captured under different weather conditions (sunny, cloudy, foggy, and rainy), with 500 frames for each weather type. We conduct a zero-shot generalization test on this dataset. %, as it closely resembles real-world application conditions in autonomous driving. 

\begin{figure}[t]
	\centering
	\includegraphics[width=\linewidth]{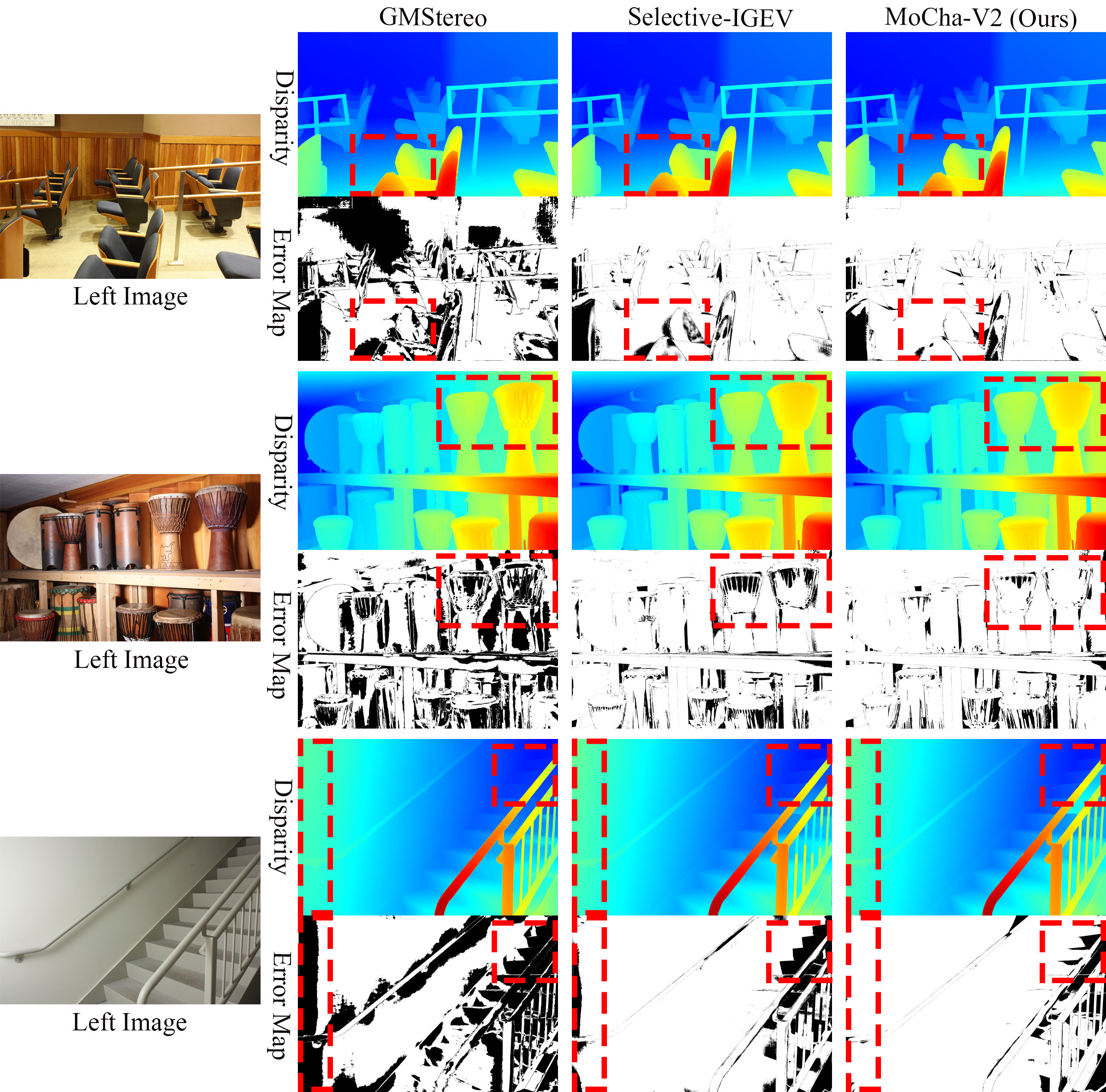}
	\caption{Visual comparisons with state-of-the-art stereo methods \cite{gmstereo,selective} on the Middlebury test set. We present the disparity map and error map with a 1-pixel threshold. In the error map, the black regions indicate areas of misestimation. }
	\label{mid_pic}
\end{figure}

\begin{table*}[t]
	\centering
	\caption{Results on the KITTI 2012 Reflective leaderboard. \textbf{Bold}: best performance, \underline{underline}: second best.}
	\label{reflective2012}
	\begin{tabular}{l|cccc|cccc|ccc}
		\toprule[1.5pt]
		\multirow{2}{*}{Method} & & \multicolumn{2}{c}{D1\textgreater2px (\%) $\downarrow$} &  &  & \multicolumn{2}{c}{D1\textgreater3px (\%) $\downarrow$} &  &  & \multicolumn{2}{c}{D1\textgreater4px (\%) $\downarrow$} \\ \cline{3-4} \cline{7-8} \cline{11-12} 
		&   & Noc & All &  &  & Noc & All &  &  & Noc & All \\ \midrule
		DPCTF-S \cite{dpctf} &  & 9.92 & 12.30 &  &  & 6.12 & 7.81 &  &  & 4.47 & 5.70 \\
		CREStereo \cite{cre}  &  & 9.71 & 11.26 &  &  & 6.27 & 7.27 &  &  & 4.93 & 5.55 \\
		NLCA-Net V2 \cite{nlca}  &  & 11.20 & 13.19 &  &  & 6.17 & 7.65 &  &  & 4.06 & 5.19 \\
		UCFNet \cite{ucfnet} &  & 9.78 & 11.67 &  &  & 5.83 & 7.12 &  &  & 4.15 & 4.99 \\
		IGEV-Stereo \cite{igev}&  & 7.29 & 8.48 &  &  & 4.11 & 4.76 &  &  & 2.92 & 3.35 \\
		RiskMin \cite{riskmin}&  & 7.57 & 9.60 &  &  & 4.11 & 5.51 &  &  & 2.87 & 3.74 \\
		GANet+ADL \cite{adl}&  & 8.57 & 10.42 &  &  & 4.84 & 6.10 &  &  & 3.43 & 4.39 \\
		Selective-IGEV \cite{selective}  &  & 6.73 & 7.84 &  &  & 3.79 & 4.38 &  &  & 2.66 & 3.05 \\	
		%HART \cite{hart}&  & 6.18 & 7.38 &  &  & 3.14 & 3.92 &  &  & 1.99 & 2.49 \\	
		DEFOM-Stereo \cite{defom} &  & 5.76 & \underline{6.72} &  &  & 3.04 & 3.56 &  &  & 1.93& \underline{2.27} \\
		AD-Stereo \cite{wang2025adstereo}  &  & 8.38 & 9.67 &  &  & 5.10 & 5.98 &  &  & 3.89 & 4.55 \\	
		IGEV++ \cite{igev++}  &  & \underline{5.68} & \textbf{6.66} &  &  & 3.71 & 4.35 &  &  & 2.60 & 3.00 \\
		MonSter++ \cite{monster++} &  & 6.47 & 7.64 &  &  & \underline{2.91} & \underline{3.45} &  &  & \underline{1.92} & 2.29 \\	
		GGEV \cite{ggev}&  & 7.33 & 9.27 &  &  & 4.04 & 5.34 &  &  & - & - \\	
		\midrule
		MoCha-Stereo (conference version) \cite{mocha}&  & 6.97 & 8.10 &  &  & 3.83 & 4.50 &  &  & 2.62 & 3.80 \\
		MoCha-V2 (Ours) &  & \textbf{5.57} & \textbf{6.66} &  &  & \textbf{2.77} & \textbf{3.38} &  &  & \textbf{1.81} & \textbf{2.11} \\
		\bottomrule[1.5pt]
	\end{tabular}
\end{table*}
\begin{figure*}[t]
\centering
\includegraphics[width=\linewidth]{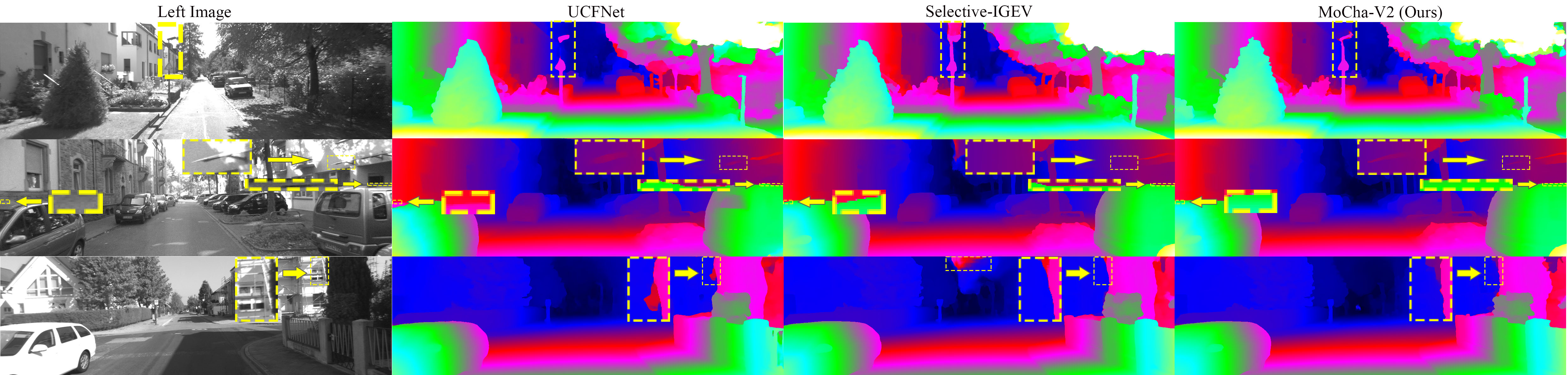}
\caption{Visual comparisons with state-of-the-art stereo methods \cite{ucfnet,selective} on the KITTI 2012 test set. In the first row of images, the streetlight presents a matching challenge due to its thin structure. UCFNet fails to capture the presence of the poles, and Selective-IGEV inaccurately estimates the sky above the streetlight as being closer in depth to the streetlight. Only MoCha-V2 provides a more accurate estimation of the streetlight's edges. In the second row, the roofs of the two cars in the image reflect natural light, which affects the accuracy of stereo matching. Both UCFNet and Selective-IGEV are impacted by this reflection and fail to accurately capture the cars' edge textures. Additionally, Selective-IGEV does not estimate the contours of the poles on the wall. In the third row, due to lighting influences, UCFNet and Selective-IGEV misinterpret the buildings in the background as part of the trees in the foreground. In contrast, MoCha-V2 delineates a more accurate edge texture.}
\label{kitti2012reflective_show}
\end{figure*}

\subsection{Implementation Details}
\label{4.1}
MoCha-V2 is implemented using the PyTorch framework and utilizes NVIDIA Tesla A100 GPUs. For all training and ablation experiments, our model employs the AdamW \cite{adamw} optimizer and clip gradients to the range [-1, 1]. We employ $22$ update iterations and a one-cycle learning rate schedule with a learning rate of $2e-4$. The training and testing settings are consistent with \cite{mocha,cre}. 

Our training strategy follows the mainstream two-stage protocol adopted by recent SOTA methods \cite{selective,gmstereo,cre}. For fair comparison, all baseline methods are evaluated using the best reported results or official pretrained models from their respective papers, and are evaluated under their original protocols. The compared SOTA methods \cite{selective,gmstereo,cre} are also trained with the same two-stage protocol, ensuring consistent training and evaluation settings across all methods.

On the Scene Flow \cite{sceneflow} dataset, MoCha-V2 is trained for $200,000$ steps with a batch size of $8$. Input images are randomly cropped to a resolution of 320 $\times$ 736 pixels. We employ data augmentation techniques such as asymmetric chromatic augmentations and spatial transformations. 

To adapt the model to Middlebury \cite{middlebury}, following existing methods \cite{selective,gmstereo,cre}, we start by fine-tuning the Scene Flow \cite{sceneflow} pretrained model with a crop size of 384 $\times$ 512 for 200,000 steps, incorporating the mixed Tartan Air \cite{tartanair}, CREStereo \cite{cre}, Scene Flow \cite{sceneflow}, Falling Things \cite{falling}, InStereo2k \cite{instereo2k}, CARLA \cite{carla}, and Middlebury datasets \cite{middlebury}. We subsequently refine the model on the mixed CREStereo \cite{cre}, Falling Things \cite{falling}, InStereo2k \cite{instereo2k}, CARLA \cite{carla}, and Middlebury datasets \cite{middlebury} using a batch size of 8 for an additional 100,000 steps. 

For training on the KITTI 2012 \cite{kitti2012} and KITTI 2015 \cite{kitti2015} datasets, we fine-tune the Scene Flow pretrained model on the combined KITTI 2012 and KITTI 2015 datasets \cite{igev} for a duration of 50,000 steps. 

For training on the ETH3D dataset \cite{eth3d}, the crop size is consistently set to 384 $\times$ 512 pixels. We first fine-tune the Scene Flow pretrained model on the mixed datasets of Tartan Air \cite{tartanair}, CREStereo \cite{cre}, Scene Flow \cite{sceneflow}, Sintel Stereo \cite{sintel}, InStereo2k \cite{instereo2k}, and ETH3D \cite{eth3d} for 300,000 steps. Then, we fine-tune it on the mixed CREStereo \cite{cre}, InStereo2k \cite{instereo2k}, and ETH3D datasets \cite{eth3d} for an additional 90,000 steps. 

The experimental settings are summarized in Table \ref{tab:protocol}, and the comparative experiments in Tables \ref{SF}–\ref{iter_time} all adopt consistent settings.

\begin{table*}[t]
	\centering
	\caption{Results on the KITTI 2015 and KITTI 2012 leaderboards. In the KITTI 2015 table, ``D1" means percentage of stereo disparity outliers in first frame, ``all" means percentage of outliers averaged over all ground truth pixels, ``fg" means percentage of outliers averaged only over foreground regions, ``bg" means percentage of outliers averaged only over background regions. In the KITTI 2015 table, ``All" in KITTI 2012 means ercentage of erroneous pixels in total, ``Noc" means percentage of erroneous pixels in non-occluded areas. Error threshold is 2 px for KITTI 2012. %\textbf{Bold}: best performance, \underline{underline}: second best.
	}
	\label{kittitable}
	\begin{tabular}{l|ccccc|ccc}
		\toprule[1.5pt]
		\multirow{2}{*}{Method} &  & \multicolumn{3}{c}{KITTI 2015 (\%) $\downarrow$} &  &  & \multicolumn{2}{c}{KITTI 2012 (\%) $\downarrow$} \\ \cline{3-5} \cline{8-9} 
		&  & D1-all & D1-fg & D1-bg &  &  & All  & Noc \\ \midrule
		CREStereo \cite{cre}  &  & 1.69 & 2.86 & 1.45 &  &   & 2.18 & 1.72\\
		NLCA-Net V2 \cite{nlca} &  & 1.77 & 3.56 & 1.41 &  &   & 2.34 & 1.83\\
		CroCo-Stereo \cite{croco} &  & 1.59 & 2.65 & 1.38 &  &  & - & - \\
		DLNR \cite{DLNR} &  & 1.76 & 2.59 & 1.60 &  &  & - & - \\
		IGEV-Stereo \cite{igev} &  & 1.59 & 2.67 & 1.38 &  &   & 2.17 & 1.71\\
		UCFNet \cite{ucfnet} &  & 1.86 & 3.33 & 1.57 &  &   & 2.17 & 1.67\\
		GMStereo \cite{gmstereo} &  & 1.77 & 3.14 & 1.49 &  &  & - & - \\
		%Any-RAFT \cite{any}& AAAI2024 &  & 1.70 & 3.04 & 1.44 &  &  & - & - \\
		DKT-IGEV \cite{dkt} &  & 1.72 & 3.05 & 1.46 &  &  & - & - \\
		LoS \cite{los} &  & 1.65 & 2.81 & 1.42 &  &   & 2.12 & 1.69\\
		Selective-IGEV \cite{selective} &  & 1.55 & 2.61 & 1.33 &  &   & 2.05& 1.59 \\ 
		AD-Stereo \cite{wang2025adstereo}  &  & 1.82 & 2.94 & 1.59 &  &  & 2.69 & 2.23 \\
		GREAT-IGEV \cite{great}  &  & \underline{1.50} & 2.59 & \underline{1.28} &  &  & \underline{2.00} & \textbf{1.51} \\
		IGEV++ \cite{igev++}  &  & 1.51 & \underline{2.54} & 1.31 &  &  & 2.03 & \underline{1.56} \\
		GEAStereo \cite{geastereo} &  & 1.51 & - & \textbf{1.27} &  &  & - & - \\
		GGEV \cite{ggev} &  & 1.70 & 3.28 & 1.38 &  &  & 2.17 & 1.66 \\
		\midrule
		MoCha-Stereo (conference version) \cite{mocha}  &  & 1.53 & \textbf{2.43} & 1.36 &  &  & 2.07  & 1.64\\
		MoCha-V2 (Ours)  &  & \textbf{1.49} & 2.62 & \textbf{1.27} &  &   & \textbf{1.91}& \textbf{1.51}\\ 
		\bottomrule[1.5pt]
	\end{tabular}
\end{table*}
\begin{figure*}[t]
	\centering
	\includegraphics[width=\linewidth]{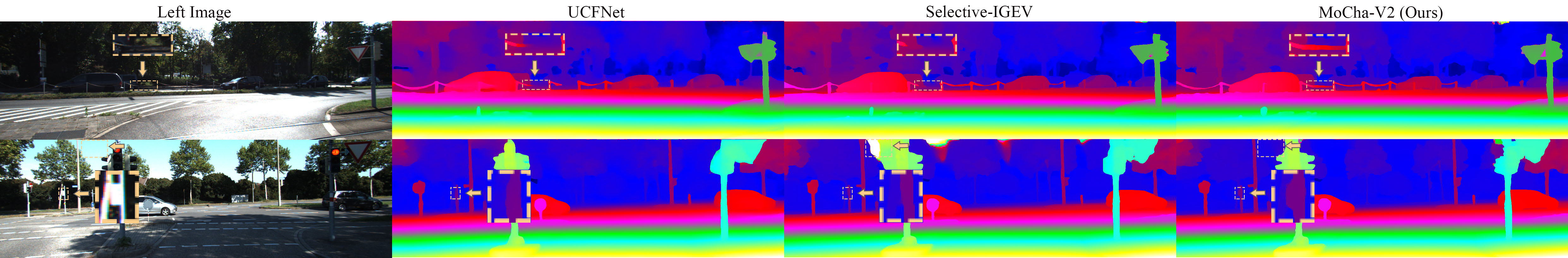}
	\caption{Visual evaluations using the KITTI 2015 test set in contrast to the SOTA techniques \cite{ucfnet,selective}. In the first row, UCFNet and Selective-IGEV failed to detect the ropes along the roadside. In the second row, Only our method successfully detects the presence of the signage.}
	\label{kitti2015_show}
\end{figure*}

\subsection{Performance Evaluation}
We contrast the SOTA techniques on Scene Flow \cite{sceneflow}, KITTI 2012 \cite{kitti2012}, KITTI 2015 \cite{kitti2015}, Middlebury \cite{middlebury}, and ETH3D \cite{eth3d} datasets. MoCha-V2 achieves well performance on each of the aforementioned datasets.

\begin{table}[t]
	\centering
	\caption{Quantitative Comparisons with SOTA Methods on the ETH3D Benchmark. Error threshold is 0.5 px.} %\textbf{Bold}: best performance, \underline{underline}: second best.}
\label{eth3d_tab}
\begin{tabular}{l|cc}
	\toprule[1.5pt]
	Method & Noc (\%) $\downarrow$ & All (\%) $\downarrow$ \\ \midrule
	RAFT-Stereo \cite{raftstereo} & 7.04 & 7.33 \\
	CREStereo \cite{cre} & 3.58 & \underline{3.75} \\
	IGEV-Stereo \cite{igev} & \underline{3.52} & 3.97 \\
	GMStereo \cite{gmstereo} & 5.94 & 6.44 \\
	CREStereo++ \cite{cre++} & 4.61 & 4.83 \\
	%Any-RAFT \cite{any} & 6.04 & 6.28 \\
	LoS \cite{los} & 3.59 & 3.83 \\
	GANet+ADL \cite{adl} & 8.36 & 8.73 \\ 
	%HART \cite{hart}  & 3.80 & 4.24 \\ 
	GGEV \cite{ggev}  & 3.70 & 4.31 \\ 
	\midrule
	MoCha-V2 (Ours)  & \textbf{3.20} & \textbf{3.68}\\ 
	\bottomrule[1.5pt]
\end{tabular}
\end{table}

\textbf{Scene Flow} \cite{sceneflow}. Among all the published approaches described in Table \ref{SF}, MoCha-V2 performs the best. MoCha-V2 achieves a new SOTA end-point error (EPE) of \textbf{0.39}, surpassing Selective-IGEV \cite{selective} by a margin of \textbf{9.09\%}.

\begin{figure}[t]
	\centering
	\includegraphics[width=\linewidth]{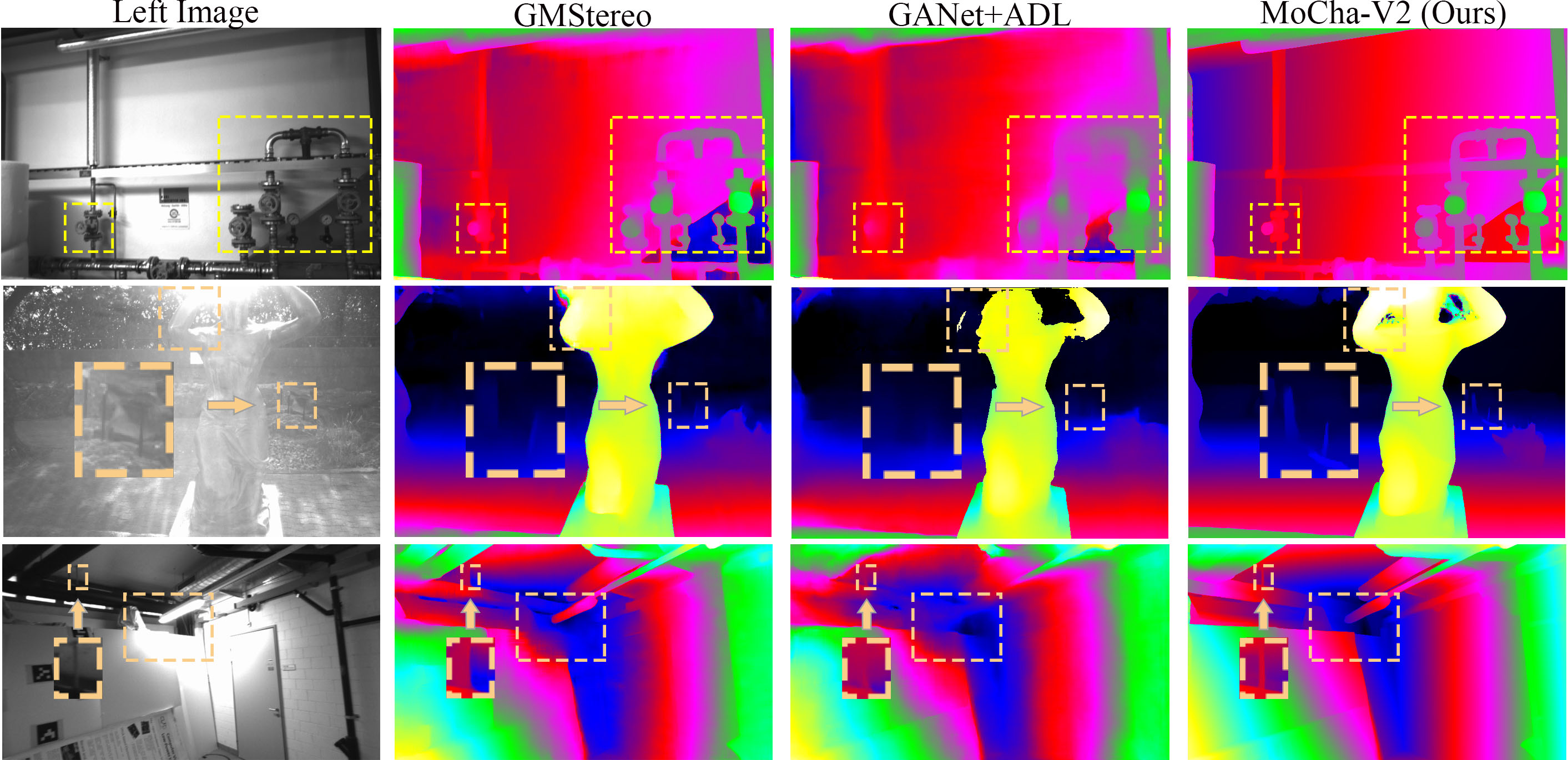}
	\caption{Visual comparisons with SOTA stereo methods \cite{gmstereo,adl} on the ETH3D test set. In the first row, GMStereo and GANet+ADL fail to capture the fine-level geometry of sluices and pipes. In the second row, reflection effects make it challenging for existing methods \cite{gmstereo,adl} to accurately identify the contours of the sculpture. Furthermore, among the three methods, only MoCha-V2 successfully detects the outline of the thin object adjacent to the sculpture. In the third row, lighting effects mislead existing methods, resulting in inaccuracies when identifying lamps, walls, and thin objects. In contrast, MoCha-V2 accurately identifies these objects and preserves their geometric contours.}
	\label{eth3d_show}
\end{figure}

\textbf{Middlebury} \cite{middlebury} imposes stringent requirements for edge detail matching. Quantitative analysis indicates that MoCha-V2 demonstrates robust matching capabilities. As shown in Table \ref{mid2014}, MoCha-V2 achieves a new SOTA of $11.4$ under the 1px error threshold, ranking \textbf{$1^{st}$} on this benchmark. The visual analysis results presented in Figure \ref{mid_pic} further demonstrate our capability to learn edge textures. As observed in the error maps, the red boxes highlight noticeably more edge-related errors in estimates from other methods. In contrast, MoCha-V2 achieves finer matching in regions with edge textures. 

\begin{table*}[t]
	\centering
	\caption{Ablation study for MoCha-V2. The baseline employed in these experiments utilized EfficientNet \cite{tan2019efficientnet} as the backbone for IGEV-Stereo \cite{igev} with 32 iterations. The Time denotes the inference time on single NVIDIA A100. ``WT" means Wavelet Transform in our MoCha-V2, ``G" means Gaussian filter in our MoCha-Stereo. The Time denotes the inference time on single NVIDIA A100.}
	\label{abl}
	\begin{tabular}{c|l|ccc|cc|c}
		\toprule[1.5pt]
		No. & Model & REMP & MCA & MCG & EPE (px) & D1\textgreater1px (\%) & Time (s) \\ \midrule
		1 & Baseline &  &  &  & 0.451 & 5.247 & 0.34 \\
		2 & MoCha-Stereo \cite{mocha} & \checkmark & \checkmark &  & 0.409 & 4.851 & 0.35 \\
		3 & MoCha-Stereo w/ WT w/o G & \checkmark & $\bigcirc$ &  & 0.401 & 4.825 & 0.35 \\
		4 & MoCha-V2 & \checkmark &  & \checkmark &  \textbf{0.381} & \textbf{4.304} & 0.32 \\ \bottomrule[1.5pt]
	\end{tabular}
\end{table*}

\textbf{KITTI 2012 Reflective} \cite{kitti2012} is specifically designed to evaluate matching performance in reflective regions. Our method demonstrates exceptional accuracy in these areas, ranking \textbf{$2^{nd}$} on the KITTI 2012 reflective benchmark at the time of submission. Evaluation results for this benchmark are presented in Table \ref{reflective2012}. 
We also provide visualizations of MoCha-V2 and compare it with existing SOTA algorithms in Figure \ref{kitti2012reflective_show}. MoCha-V2 demonstrates the capability to accurately estimate object boundaries under various conditions.

\textbf{KITTI 2015} \cite{kitti2015} and \textbf{KITTI 2012} \cite{kitti2015}. 
Our method achieves SOTA performance on both the KITTI 2012 and KITTI 2015 benchmarks. The quantitative comparisons are presented in Table \ref{kittitable} and Figure \ref{kitti2015_show}. Selective-IGEV \cite{selective} outperforms MoCha-V2 on background metrics. However, due to the greater complexity of foreground details, MoCha-V2 exceeds Selective-IGEV’s performance in the foreground by \textbf{8.05\%}, leading to superior overall results. Furthermore, MoCha-V2 achieves better results in occluded regions, demonstrating its ability to infer more accurate geometry in occluded areas based on recurrent geometric information.

\textbf{ETH3D.} MoCha-V2 outperforms the majority of published methods on the ETH3D \cite{eth3d} leaderboard. The quantitative comparisons are presented in Table \ref{eth3d_tab}. Visual comparisons are shown in Figure \ref{eth3d_show}, by effectively capturing geometric information, MoCha-V2 is able to preserve the texture structure of thin objects.

\begin{figure}[]
\centering
\includegraphics[width=\linewidth]{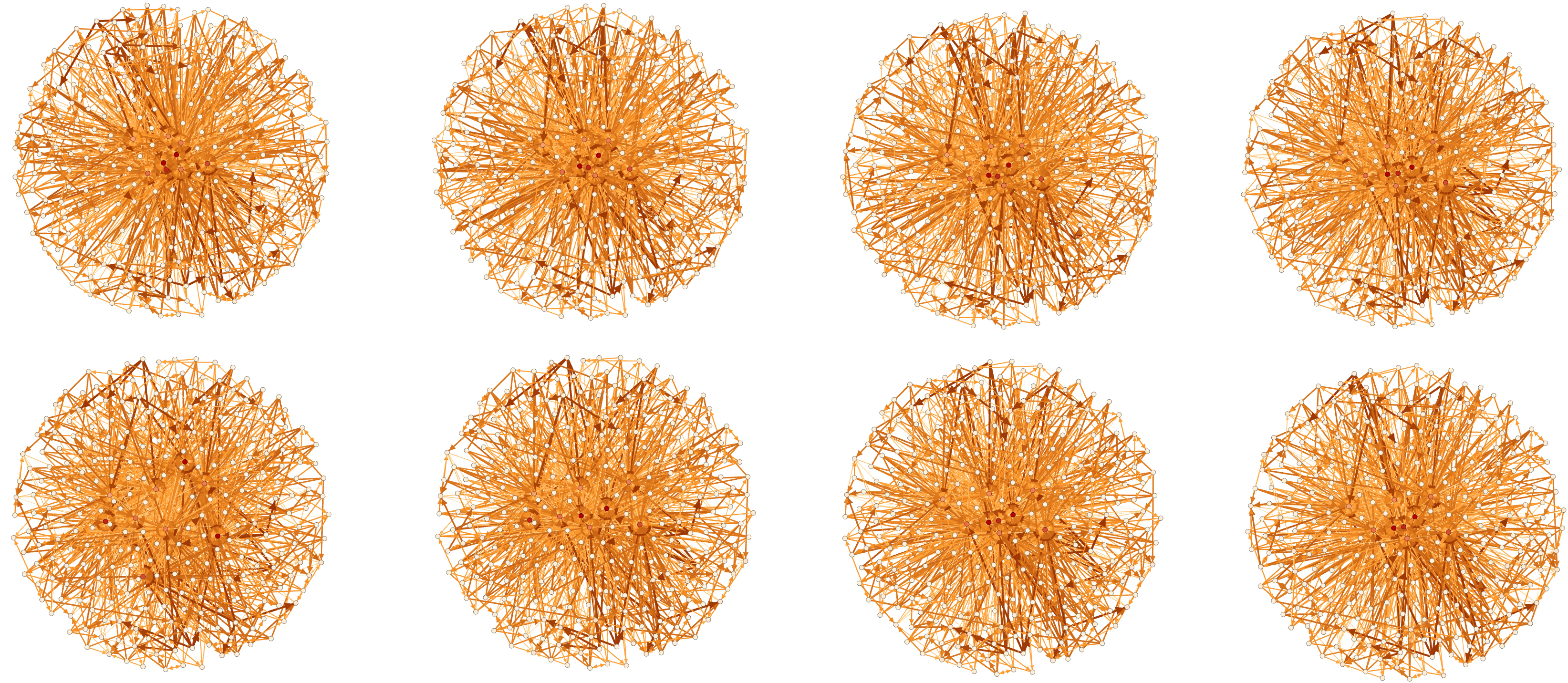}
\caption{Visualization of Motif Correlation Graphs computed from a pair of images in the Scene Flow dataset. The darker the color of a node, the greater its weight. Arrows point from each node to the node with the smallest Euclidean distance to it.
}
\label{mcg1}
\end{figure}

\begin{figure}[]
\centering
\includegraphics[width=\linewidth]{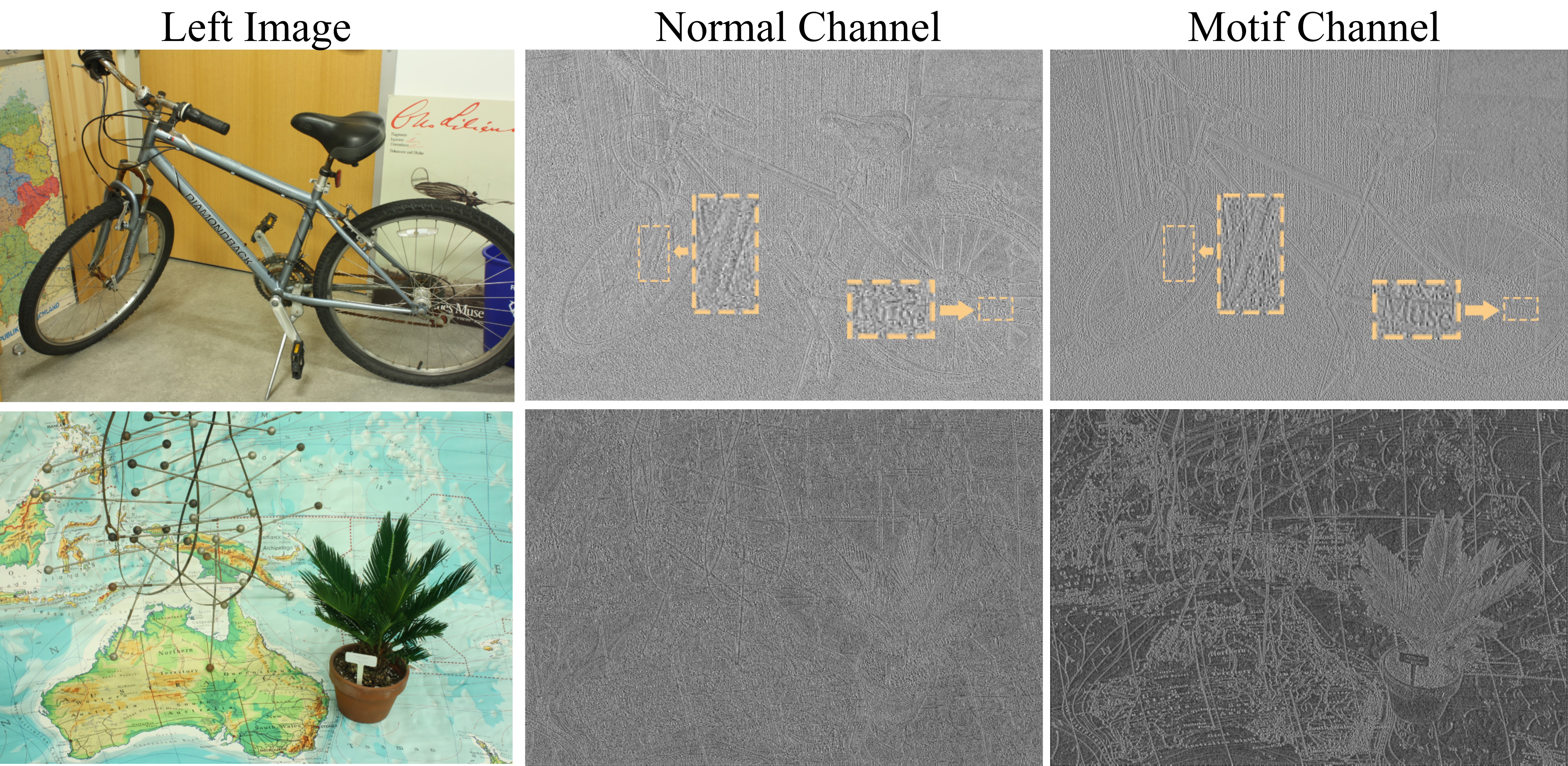}
\caption{Visualization of feature channels. We utilize Principal Component Analysis (PCA) to reduce the channel features to a one-dimensional feature map and visualize it.
}
\label{mcg2}
\end{figure}

\subsection{Ablation Study}

To validate and gain a comprehensive understanding of our model's architecture, we conducted ablation experiments. All hyperparameters are set to the same values as those used during the pretraining phase on the Scene Flow dataset.

\textbf{Effectiveness of Wavelet Transform.}
Unlike our conference version \cite{mocha}, we further utilize low-frequency information. Specifically, we employ wavelet transform (WT) in place of a Gaussian filter in MoCha-Stereo. As illustrated in Experiment No. 3 of Table \ref{abl}, incorporating recurrent low-frequency features also benefits stereo matching.

\textbf{Effectiveness of Motif Correlation Graph (MCG).}
In our conference version, Motif Channel Attention (MCA) \cite{mocha} is an adaptive learning approach for the Motif Channel. However, it lacks robustness and interpretability for practical applications such as autonomous driving. To better serve these applications, we introduce MCG in this version. 
Figure \ref{mcg1} shows the visualization of our Motif Correlation Graphs. We use Gephi to visualize our MCGs, with the layout type set to ``Fruchterman Reingold" and the following parameters: Area=100,000, Gravity=10.0. 
MCG outperforms MCA, as indicated by Experiment No. 4 in Table \ref{abl}. As shown in Figure \ref{mcg2}, the attention map generated by the MCG effectively preserves the geometric details. MCG also makes MoCha-V2 cheaper than MoCha-Stereo. MCG only requires the construction of a single motif channel for each feature channel, enabling it to achieve faster inference times compared to MCA.

\begin{table}[]
\caption{Ablation study for iterations. The metric here is EPE (px). \textbf{Bold}: Best Performance.}
\label{iter_abl}
	\begin{tabular}{l|ccccccc}
	\toprule[1.5pt]
	\multirow{2}{*}{Method} &  & \multicolumn{6}{c}{Number of Iterations} \\ \cline{3-8} 
	&  & 1 & 2 & 4 & 8 & 16 & 32 \\	\midrule
	%DLNR \cite{DLNR} &  & 1.50 & 0.88 & 0.64 & 0.52 & 0.50 & 0.48 \\
	IGEV-Stereo \cite{igev} &  & 0.66 & 0.62 & 0.55 & 0.50 & 0.47 & 0.47 \\ 
	Selective-IGEV \cite{selective} &  & 0.65 & 0.60 & 0.53 & 0.48 & 0.45 & 0.44 \\ 
	MoCha-Stereo \cite{mocha} &  & 0.56 & 0.52 & 0.46 & 0.42 & 0.41 & 0.41 \\
	\midrule
	MoCha-V2 (Ours) &  & \textbf{0.49} & \textbf{0.45} & \textbf{0.42} & \textbf{0.40} & \textbf{0.39} & \textbf{0.38} \\ \bottomrule[1.5pt]
	\end{tabular}
\end{table}

\textbf{Number of Iterations.}
MoCha-V2 demonstrates strong performance even with a reduced number of iterations. As shown in Table \ref{iter_abl}, MoCha-V2 achieves SOTA results without the need for a large number of iterations. With just 4 iterations, MoCha-V2 surpasses IGEV-Stereo \cite{igev} (EPE: 0.47 px, time: 0.37 s) by over 4.3\% in accuracy and reduces inference time by a notable \textbf{45.9\%}. This allows users to balance time efficiency and performance according to their specific application needs. 
What is more, MoCha-V2 only needs to build one motif channel for each channel, allowing it to achieve faster inference times than MoCha-Stereo \cite{mocha}. As depicted in Table \ref{iter_time}, we perform more efficiently than our conference version.

\begin{table}[]
\centering
\caption{A comparison of accuracy and speed between MoCha-Stereo and MoCha-V2. The ``+" symbol indicates the percentage by which MoCha-V2 outperforms MoCha-Stereo, while the ``-" shows the reverse. ``Iters" means the number of iterations. ``Time" refers to the inference time on a single NVIDIA Tesla A100.}
\label{iter_time}
\begin{tabular}{c|ccccc|ccc}
	\toprule[1.5pt]
	&  & \multicolumn{3}{c}{EPE (px)} &  &  & \multicolumn{2}{c}{Time (s)} \\ \cline{3-5} \cline{8-9} 
	\multirow{-2}{*}{Iters} &  & \cite{mocha} & V2 & $\pm$ &  &  & \cite{mocha} & V2 \\ \midrule
	1 &  & 0.56 & 0.49 & + \textless 1\% &  &  & 0.19 & 0.17 \\
	2 &  & 0.52 & 0.45 & + 1.9\% &  &  & 0.20 & 0.18 \\
	4 &  & 0.46 & 0.42 & + 2.2\% &  &  & 0.22 & 0.20\\
	8 &  & 0.42 & 0.40 & + \textless 1\% &  &  & 0.26 & 0.23 \\
	16 &  & 0.41 & 0.39 & + 2.4\% &  &  & 0.30 & 0.27 \\
	32 &  & 0.41 & 0.38 & + 4.9\% &  &  & 0.35 & 0.31   \\ \bottomrule[1.5pt]
\end{tabular}
\end{table}

\section{Conclusion}
This paper proposes MoCha-V2, a novel stereo matching network. Specifically, MoCha-V2 utilizes a white-box graph to focus on recurrent geometric details within the feature channel, enabling the preservation of edges during the deep learning process. 
We explicitly extract the features that play a crucial role in restoring edge details, which offers meaningful implications for stereo matching applications. 
Experiments validate the applicability of our approach to real-world scenarios. Our MoCha-V2 achieves SOTA performance on the Middlebury, KITTI, and ETH3D benchmarks and demonstrates strong generalization capability in unseen real-world scenarios.

While it is regrettable that MoCha-V2 is not a fully white-box stereo matching method, the deep learning processes outside of the Motif Correlation Graph (MCG) may still lead to safety concerns. The reason we have not yet achieved a fully white-box process is that the gap between interpretability techniques and the performance of deep learning modules is still significant, and a fully transparent approach currently struggles to achieve higher accuracy. We plan to further enhance the interpretability of learning-based stereo matching methods in the future.

\balance

\section*{Acknowledgment}

This work was supported in part by the Science and Technology Planning Project of Guizhou Province, Department of Science and Technology of Guizhou Province, China under Grant QianKeHe [2024] Key 001; in part by the Science and Technology Planning Project of Guizhou Province, Department of Science and Technology of Guizhou Province, China under Grant [2023]159; and in part by the Natural Science Research Project of Guizhou Provincial Department of Education, China under Grant QianJiaoJi[2022] 029, QianJiaoHeKY[2021]022.
\iffalse
\section{A Final Checklist}
\begin{enumerate}{}{}
\item{Make sure that your equations are numbered sequentially and there are no equation numbers missing or duplicated. Avoid hyphens and periods in your equation numbering. Stay with IEEE style, i.e., (1), (2), (3) or for sub-equations (1a), (1b). For equations in the appendix (A1), (A2), etc.}. 
\item{Are your equations properly formatted? Text, functions, alignment points in cases and arrays, etc. }
\item{Make sure all graphics are included.}
\item{Make sure your references are included either in your main LaTeX file or a separate .bib file if calling the external file.}
\end{enumerate}
\fi

\bibliographystyle{IEEEtran}
\bibliography{IEEEabrv,mybibfile}


\begin{IEEEbiography}[{\includegraphics[width=1in,height=1.25in,clip,keepaspectratio]{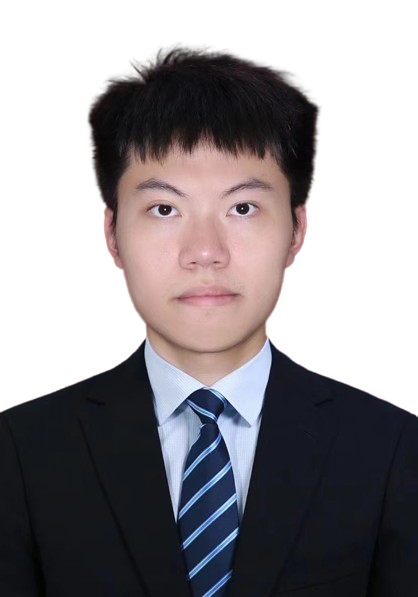}}]{Ziyang Chen}
	is currently a Ph.D. student at the School of Informatics, Xiamen University, China. He obtained his Master’s degree from the College of Computer Science and Technology, Guizhou University, under the supervision of Dr. Yongjun Zhang. He has published several papers in journals and conferences, including \textit{IEEE Conference on Computer Vision and Pattern Recognition (CVPR)}, and \textit{IEEE Journal of Selected Topics in Applied Earth Observations and Remote Sensing}. His research interests lie in 3D vision, stereo matching, attention mechanism, remote sensing, and intelligent transportation.
\end{IEEEbiography}

\begin{IEEEbiography}[{\includegraphics[width=1in,height=1.25in,clip,keepaspectratio]{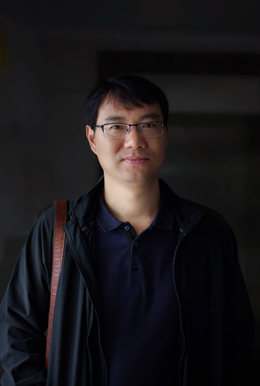}}]{Yongjun Zhang}
	(Member, IEEE) received the M.Sc. and PhD degrees in software engineering from Guizhou University, Guiyang, China in 2010 and 2015 respectively. From 2012 to 2015, he is a joint training doctoral student of Peking University, Beijing, China, and Guizhou University, studying in the key laboratory of integrated microsystems of Peking University Shenzhen Graduate School, Shenzhen, China.
	
	He is currently an associate professor with Guizhou University. His research interests include the intelligent image algorithms of computer vision, such as scene target detection, remote sensing, stereo matching and low-level vision.
\end{IEEEbiography}

\begin{IEEEbiography}[{\includegraphics[width=1in,height=1.25in,clip,keepaspectratio]{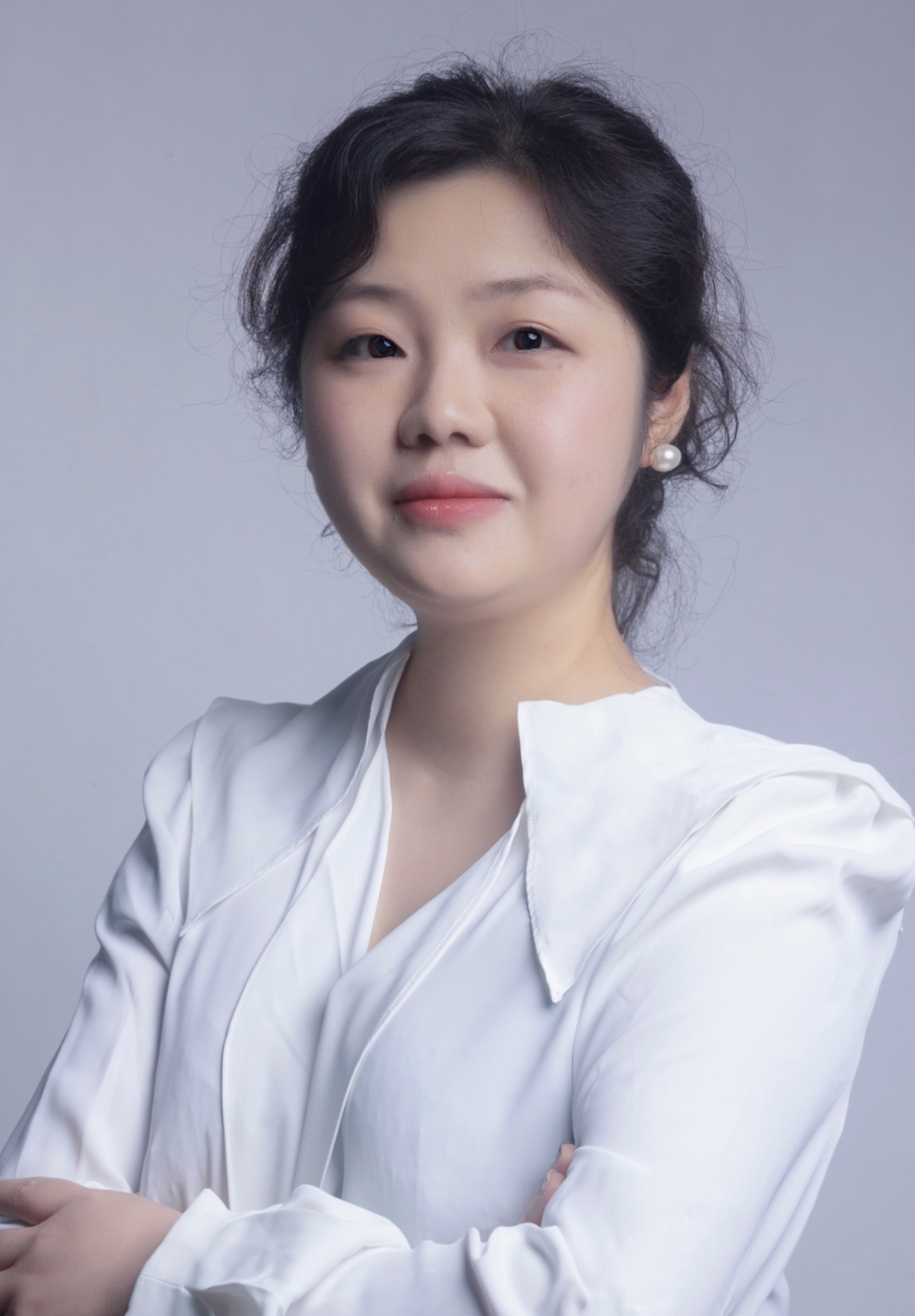}}]{Wenting Li}
	received the B.Sc. degree in computer science and technology from Zhengzhou University of Aeronautics, Zhengzhou, China, in 2006, the M.Sc. degree in computer science from GuiZhou University, Guiyang, China, in 2010, and the Ph.D. degree in computer technology and application from the Macau University of Science and Technology, Macau, China, in 2017.
	
	She is a professor with the Guizhou University of Commerce, Guiyang, China, since 2018. Her research interests include intelligent transportation, computer vision, and data analysis.
\end{IEEEbiography}

\begin{IEEEbiography}[{\includegraphics[width=1in,height=1.25in,clip,keepaspectratio]{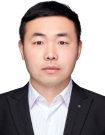}}]{Bingshu Wang}
	received the B.S. degree in computer science and technology from Guizhou University, 	Guiyang, China, in 2013, the M.S. degree in electronic science and technology (integrated circuit	system) from Peking University, Beijing, China,	in 2016, and the Ph.D. degree in computer science from the University of Macau, Macau, in 2020. He is currently an Associate Professor with the School of Software, Northwestern Polytechnical	University, Suzhou, China. His current research	interests include computer vision, intelligent video
	analysis, and machine learning.	
\end{IEEEbiography}

\begin{IEEEbiography}[{\includegraphics[width=1in,height=1.25in,clip,keepaspectratio]{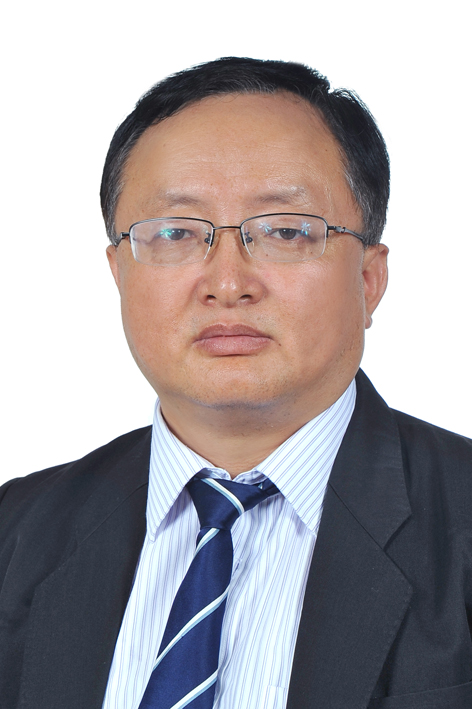}}]{Yong Zhao}
	(Member, IEEE) received the Ph.D. degree in automatic control and applications from 	Southeast University, Nanjing, China, 1991.	He then joined Zhejiang University, Hangzhou,
	China, as an Assistant Researcher. In 1997, he went	to Concordia University, Montreal, QC, Canada, as a Post-Doctoral Fellow. 
	
	He was a Senior Audio/Video Compression Engineer with Honeywell Corporation, Mississauga, ON, Canada, in May 2000. In 2004, he became an Associate Professor at the Peking University Shenzhen Graduate School, Shenzhen, China. In 2025, he became a Researcher at the School of Intelligent Manufacturing and Future Technology, Fujian Fuyao University of Science and Technology, Fuzhou, China, where he is currently the Header of the lab Mobile Video Networking Technologies. He is working on computer vision, machine learning, video analytics, and video compression with a special focus on applications of these new theories and technologies to various industries. His team has developed many innovative products and projects that have been successful in the market.
\end{IEEEbiography}

\begin{IEEEbiography}[{\includegraphics[width=1in,height=1.25in,clip,keepaspectratio]{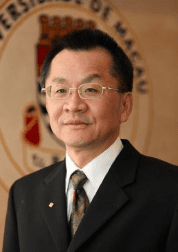}}]{C. L. Philip Chen}
	(Life Fellow, IEEE) was a recipient of the 2016 Outstanding Electrical and Computer Engineers Award from his alma mater, Purdue University (in 1988), after he graduated from the University of Michigan at Ann Arbor, Ann Arbor, MI, USA in 1985.
	
	He is currently the Chair Professor and Dean of the College of Computer Science and Engineering, South China University of Technology.  He is a Fellow of IEEE, AAAS, IAPR, CAA, and HKIE; a member of Academia Europaea (AE), and a member of European Academy of Sciences and Arts (EASA). He received IEEE Norbert Wiener Award in 2018 for his contribution in systems and cybernetics, and machine learnings, received two times best transactions paper award from \textit{IEEE Transactions on Neural Networks and Learning Systems} for his papers in 2014 and 2018 and he is a highly cited researcher by Clarivate Analytics from 2018-2022. His current research interests include cybernetics, systems, and computational intelligence. He was the Editor-in-Chief of the \textit{IEEE Transactions on Cybernetics}, the Editor-in-Chief of the \textit{IEEE Transactions on Systems, Man, and Cybernetics: Systems}, President of \textit{IEEE Systems, Man, and Cybernetics Society}.
\end{IEEEbiography}

\end{document}